\pdfoutput=1
\documentclass[10pt,twocolumn,letterpaper]{article}

\usepackage[pagenumbers]{cvpr} %

\usepackage{lipsum}
\usepackage{array}
\usepackage{times}
\usepackage{epsfig}
\usepackage{graphicx}
\usepackage{float}
\usepackage{wrapfig}
\usepackage{amsmath,amssymb,amsthm}
\usepackage{algorithm,algorithmicx,algpseudocode}
\usepackage{bm,xspace}
\usepackage{comment}
\usepackage{multirow}
\usepackage{balance}
\usepackage{url}
\usepackage{booktabs}
\usepackage{etoolbox,siunitx}
\usepackage{calc}
\usepackage{pifont,hologo}
\usepackage{color}
\usepackage{adjustbox}
\usepackage{amsmath}
\usepackage{enumitem}
\usepackage{bbding}  %
\usepackage[normalem]{ulem}  %
\usepackage{contour}
\usepackage[table]{xcolor}
\usepackage{soul}%
\usepackage{tocloft} %
\usepackage{etoc} %
\PassOptionsToPackage{dvipsnames}{xcolor}
\usepackage[pagebackref,breaklinks,colorlinks,linkcolor=link,citecolor=cite]{hyperref}
\usepackage{natbib}
\PassOptionsToPackage{square,numbers,comma,sort}{natbib}

\newcommand{\authorskip}{\hspace{0.4mm}}
\newcommand{\institutionskip}{\hspace{5.0mm}}

\definecolor{blue}{HTML}{004bb3}
\definecolor{red}{HTML}{cc1100}
\definecolor{orange}{HTML}{cc7700}
\definecolor{gray}{HTML}{efefef}
\definecolor{darkgreen}{HTML}{228B22}
\definecolor{darkgray}{HTML}{757575}
\definecolor{cite}{HTML}{3270b5}
\definecolor{link}{HTML}{cc1100}
\definecolor{scratch}{HTML}{001219}
\definecolor{pretrain}{HTML}{0A9396}

\newcommand{\scratch}{\textcolor{scratch}{$\mathbf{\circ}$\,}}
\newcommand{\pretrain}{\textcolor{pretrain}{$\bullet$\,}}

\contourlength{0.8pt}

\newcommand{\figref}[1]{Fig.~\ref{#1}}
\newcommand{\tabref}[1]{Tab.~\ref{#1}}
\newcommand{\secref}[1]{Sec.~\ref{#1}}
\renewcommand{\eqref}[1]{Eq.~\ref{#1}}

\newcolumntype{x}[1]{>{\centering\arraybackslash}p{#1}}
\newcolumntype{y}[1]{>{\raggedright\arraybackslash}p{#1}}
\newcolumntype{z}[1]{>{\raggedleft\arraybackslash}p{#1}}
\newcommand{\tablestyle}[2]{\setlength{\tabcolsep}{#1}\renewcommand{\arraystretch}{#2}\centering\footnotesize}

\DeclareMathSymbol{@}{\mathord}{letters}{"3B}

\newcommand\mypara[1]{\vspace{0mm}\noindent\textbf{#1}}

\makeatletter
\DeclareRobustCommand\onedot{\futurelet\@let@token\@onedot}
\def\@onedot{\ifx\@let@token.\else.\null\fi\xspace}
\def\eg{\emph{e.g}\onedot} 
\def\ie{\emph{i.e}\onedot}

\newcommand*{\Rom}[1]{\expandafter\@slowromancap\romannumeral #1@}
\newcommand*{\rom}[1]{\expandafter\romannumeral #1}

\def\1{\bm{1}}

\def\vc{{\bm{c}}}

\def\vx{{\bm{x}}}
\def\vy{{\bm{y}}}

\DeclareMathAlphabet{\mathsfit}{\encodingdefault}{\sfdefault}{m}{sl}
\SetMathAlphabet{\mathsfit}{bold}{\encodingdefault}{\sfdefault}{bx}{n}

\def\sC{{\mathcal{C}}}
\def\sD{{\mathcal{D}}}

\let\originalleft\left
\let\originalright\right
\renewcommand{\left}{\mathopen{}\mathclose\bgroup\originalleft}
\renewcommand{\right}{\aftergroup\egroup\originalright}

\def\paperID{199}
\def\confName{CVPR}
\def\confYear{2024}

\begin{document}

\title{Towards Large-scale 3D Representation Learning \\ with Multi-dataset \textcolor{blue}{P}oint \textcolor{blue}{P}rompt \textcolor{blue}{T}raining}
\author{
Xiaoyang Wu\textsuperscript{\mdseries1} \authorskip 
Zhuotao Tian\textsuperscript{\mdseries2} \authorskip 
Xin Wen\textsuperscript{\mdseries1} \authorskip 
Bohao Peng\textsuperscript{\mdseries2} \authorskip 
Xihui Liu\textsuperscript{1} \authorskip
Kaicheng Yu\textsuperscript{3,4} \authorskip
Hengshuang Zhao\textsuperscript{1*}
\vspace{2mm}
 \\ 
\textsuperscript{1}The Univeristy of Hong Kong\institutionskip 
\textsuperscript{2}The Chinese Univeristy of Hong Kong\\
\textsuperscript{3}Westlake University\institutionskip 
\textsuperscript{4}Alibaba Group \\ 
{\tt\small \url{https://github.com/Pointcept/Pointcept}}
}

\twocolumn[{%
\renewcommand\twocolumn[1][]{#1}%
\vspace{-15mm}
\maketitle
\vspace{-10mm}
\begin{center}
    \centering
    \captionsetup{type=figure}
    \includegraphics[width=\linewidth]{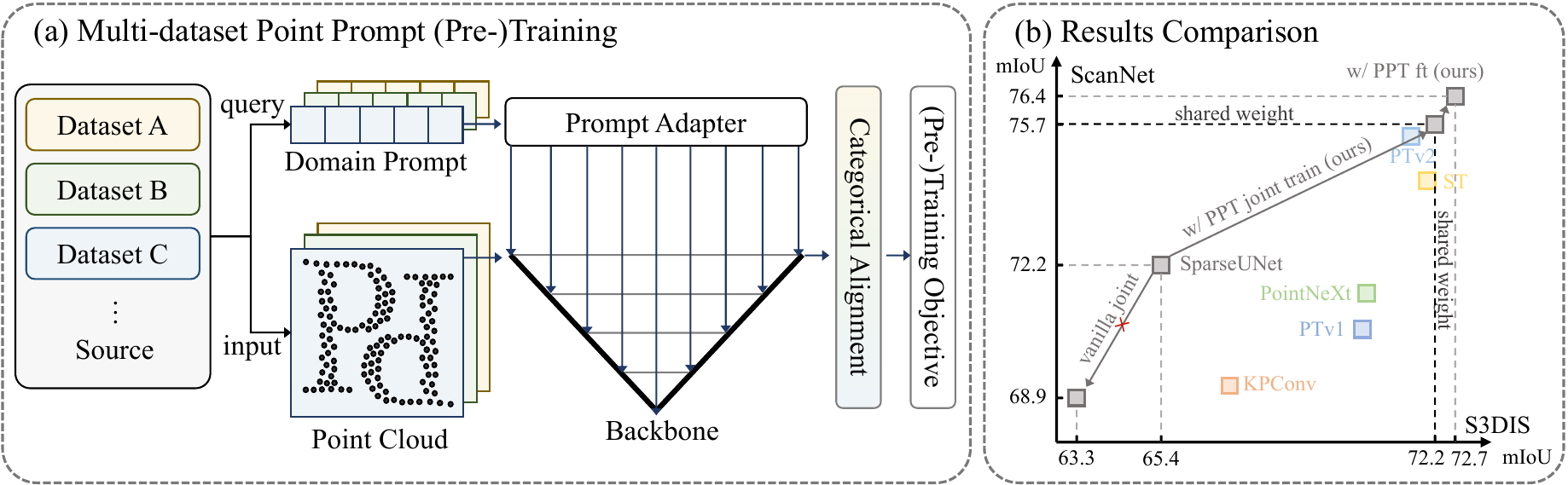}
    \vspace{-4mm}
    \captionof{figure}{\textbf{Multi-dataset synergistic training with Point Prompt Training~(PPT).} (a) Our \textit{PPT Framework} is comprised of two key components: 1. The domain prompt adapter adapts the backbone to various dataset-specific contexts with a set of domain-specific prompts; 2. The categorical alignment process empowers the model to effectively undergo training within multiple category spaces concurrently in the supervised setting. (b) The \textit{Result Comparison} plot reveals that PPT delivers state-of-the-art performance across both datasets only with one single shared-weight backbone, and fine-tuning on any single specific dataset can further enhance the results.}\label{fig:teaser}
\end{center}%
}]
\renewcommand{\thefootnote}{\fnsymbol{footnote}}
\footnotetext[1]{Corresponding author.}
\begin{abstract}
\vspace{-4mm}
The rapid advancement of deep learning models is often attributed to their ability to leverage massive training data. In contrast, such privilege has not yet fully benefited 3D deep learning, mainly due to the limited availability of large-scale 3D datasets. Merging multiple available data sources and letting them collaboratively train a single model is a potential solution. However, due to the large domain gap between 3D point cloud datasets, such mixed supervision could adversely affect the model's performance and lead to degenerated performance (\ie, \textit{negative transfer}) compared to single-dataset training. In view of this challenge, we introduce \textit{Point Prompt Training} (PPT), a novel framework for multi-dataset synergistic learning in the context of 3D representation learning that supports multiple pre-training paradigms. Based on this framework, we propose Prompt-driven Normalization, which adapts the model to different datasets with domain-specific prompts and Language-guided Categorical Alignment that decently unifies the multiple-dataset label spaces by leveraging the relationship between label text. Extensive experiments verify that PPT can overcome the negative transfer associated with synergistic learning and produce generalizable representations. Notably, it achieves state-of-the-art performance on each dataset using a single weight-shared model with supervised multi-dataset training. Moreover, when served as a pre-training framework, it outperforms other pre-training approaches regarding representation quality and attains remarkable state-of-the-art performance across over ten diverse downstream tasks spanning both indoor and outdoor 3D scenarios.
\end{abstract}

\vspace{-6mm}
\section{Introduction}
\label{sec:intro}

The rapid advancement of deep learning models in various domains, \eg, 2D vision~\cite{tian2021divide,goyal2021self,wang2022painter,kirillov2023segany} and natural language processing~\cite{kaplan2020scaling,aribandi2022ext,touvron2023llama,openai2023gpt4}, are often attributed to the availability of massive training data, which enable them to learn rich and discriminative representations and generalize well to a wide spectrum of downstream applications. Such privilege, in contrast, has not yet fully benefited 3D vision, primarily due to two challenges: previous representation learning frameworks exhibit constraints in processing larger-scale point cloud data efficiently (\ie, they build on raw frames rather than the scene-level point cloud~\cite{xie2020pointcontrast,hou2021exploring}), and current 3D datasets are often limited in scale (\eg, the commonly used ScanNet~\cite{dai2017scannet} only contains 1.6K scans, while image datasets are often at million scale~\cite{deng2009imagenet,sharma2018conceptual}). As a complement to one recent work~\cite{wu2023masked} which explores the first problem, we tackle the second challenge: \textit{scaling up 3D representation learning with limited data in separated domains.}

A potential approach to circumvent the data scarcity issue is to merge multiple available data sources and train on them collaboratively (termed multi-dataset synergistic training) to supervise a single model, which is expected to leverage the information from all sources and learn more generalizable representations. However, large domain gaps exhibit between 3D datasets, and directly combining multiple data sources can lead to \textit{negative transfer}, a phenomenon where differences in data distribution among the sources adversely affect the model's performance. As shown in \tabref{tab:pilot_transfer}, naively joint training with merged data (ScanNet~\cite{dai2017scannet}, S3DIS~\cite{armeni2016s3dis}, and Structured 3D~\cite{zheng2020structured3d}) leads to degenerated performance on the target dataset. In other words, leveraging additional training data from other datasets could be harmful. Though similar problems have been studied in 2D scene understanding~\cite{kim2022learning,wang2022cross,zhou2022simple,yao2023detclipv2,vandenhende2021multi}, the large domain gap between 3D datasets, and their sparse and heavily long-tailed nature makes it a much harder task that requires non-trivial solutions. \looseness=-1

To tackle the challenge, we present a novel framework, termed \textit{Point Prompt Training}~(PPT), specifically designed for multi-dataset synergistic training within the 3D representation learning context (see \figref{fig:teaser}\textcolor{red}{a}).
Unlike the 2D counterparts that adopt prompt learning to adapt pre-trained models to specific \textit{downstream tasks}~\cite{jia2022vpt,zhou2022coop,zang2022unified,ju2022prompting}, our framework tackles \textit{pre-training} directly. Moreover, the proposed framework is universal, supporting both supervised and unsupervised pre-training, and evaluation on the target dataset could be done either directly (if the target dataset is included in supervised pre-training) or via transfer learning. 

Based on this framework, we explore multi-dataset synergistic training for 3D representation learning from two perspectives: learning a \textit{domain prompt adapter} that allows the network to model the intrinsic variance within different data sources while maintaining optimal generalizable representations and forming a \textit{unified label space} that avoids inconsistency in categorical supervision and allows aligned guidance between datasets. Multiple design options are investigated, and we adopt the \textit{Prompt-driven Normalization} and \textit{Language-guided Categorical Alignment} as our final strategies. \looseness=-1

The effectiveness of PPT is demonstrated through extensive experiments, which show that our proposed method can overcome the negative transfer associated with synergistic learning and produce generalizable representations. Notably, PPT attains state-of-the-art performance across various benchmarks, including ScanNet~\cite{dai2017scannet} and S3DIS~\cite{armeni2016s3dis}, using a shared-weight model trained on multiple indoor datasets. Additionally, it achieves comparable state-of-the-art results on SemanticKITTI~\cite{behley2019semantickitti}, nuScenes~\cite{caesar2020nuscenes}, and Waymo~\cite{sun2020waymo} using a shared-weight model trained on diverse outdoor datasets. Furthermore, as a pre-training strategy, PPT outperforms other techniques in terms of representation quality, demonstrating superior performance across an array of tasks encompassing both indoor and outdoor scenarios (partially in \figref{fig:teaser}\textcolor{red}{b}).

In conclusion, as an effort toward large-scale 3D representation learning, this work introduces the multi-dataset synergistic training setting, points out the negative transfer issue in naive baselines, and presents a unified point prompt training framework that addresses this problem with Prompt-driven Normalization and Language-guided Categorical Alignment.

\section{Multi-dataset Synergistic Training}
\label{sec:multi-dataset}
In this section, we briefly demonstrate the setting (\secref{sec:setup}) in multi-dataset synergistic training for 3D representation learning and uncover the challenges in this setup through a pilot study (\secref{sec:pilot}).

\subsection{Problem Setup}
\label{sec:setup}
\mypara{Training objective.}
In the context of supervised multi-dataset synergistic learning, the objective is to learn a single model capable of effectively performing downstream tasks on multiple datasets. Specifically, denote each dataset as $\sD_i = \{(\vx_j^i, \vy_j^i)\}$, where $1 \leq i \leq n$, $n$ stands for the number of datasets, and $(\vx_j^i, \vy_j^i)$ represents data-label pairs that construct a dataset. Our goal is to train a model $f(\vx;\theta)$ parameterized by $\theta$, such that the cumulative loss across all datasets is minimized:
\begin{align}
    \mathop{\operatorname{argmin}}_{\theta} \sum_{i=1}^n \frac{1}{|\sD_i|} \sum_{\left(\vx_j^i, \vy_j^i\right) \in \sD_i} \mathcal{L}\left(f\left(\vx_j^i; \theta\right), \vy_j^i\right) \,,
    \label{eq:multi-dataset-training}
\end{align}
where $\mathcal{L}$ denotes the sample-wise loss function. Besides, substituting the supervised loss function with an unsupervised objective allows for reformulation in the context of unsupervised learning.

\begin{table*}[t!]
    \begin{minipage}{1\textwidth}
     \centering
        \caption{\textbf{Datasets summary and joint training transfer among tasks.} The entry at row $i$ and column $j$ indicates the semantic segmentation mIoU on dataset $i$ using a SparseUNet~\cite{choy20194d, spconv2022} joint-trained on datasets $i$ and $j$. The \textit{All} column represents combining all data sources. The diagonal elements represent using only the original dataset $i$. Note that Structured3D is originally panoramic images, and we converted it to point cloud data following Swin3D~\cite{yang2023swin3d}.
        Moreover, we compute the sampling ratio based on each dataset's best performance necessary iteration number. The effects of different sampling strategies are further explored in the ablation study (\secref{sec:ablation}) and Appendix.}
        \label{tab:pilot_transfer}
        \vspace{-2mm}
        \tablestyle{8pt}{1.1}
        \begin{tabular}{llllrccccc}
\toprule
 & \multicolumn{4}{c}{Dataset details} & \multicolumn{4}{c}{Baseline results w/ diff. joint training data} &  \multicolumn{1}{c}{\multirow{2}{*}{\begin{tabular}[c]{@{}c@{}} Ours\\ (All)\end{tabular}}}\\
\cmidrule(lr){2-5} \cmidrule(lr){6-9}
Target data&Source &Sparsity &Complexity &Scans &ScanNet &S3DIS &Struct.3D &All & \\
\midrule
ScanNet &Real &Sparse &Large rooms &1613 &\underline{72.2} &71.8 &65.9 &68.9$\,_\text{\textcolor{red}{(-3.3)}}$ &\textbf{75.7}$\,_\text{\textcolor{darkgreen}{(+3.5)}}$ \\
S3DIS &Real &Dense &School office &272 &64.1 &\underline{65.4} &62.8 &63.3$\,_\text{\textcolor{red}{(-2.1)}}$ &\textbf{72.2}$\,_\text{\textcolor{darkgreen}{(+6.8)}}$ \\
Struct.3D &Synth. &Dense &Suite &3500 &73.7 &74.2 &\underline{74.5} &72.9$\,_\text{\textcolor{red}{(-1.6)}}$ &\textbf{75.8}$\,_\text{\textcolor{darkgreen}{(+1.3)}}$ \\
\bottomrule
\end{tabular}
    \end{minipage}
    \vspace{-4mm}
\end{table*}

\mypara{Task.} The nature of 3D scene understanding has a higher level of complexity and richer contextual information~\cite{xie2020pointcontrast, hou2021exploring}, which requests a challenging and versatile task for developing and evaluating advanced learning techniques. Specifically, we mainly target scene-level \textit{semantic segmentation} for supervised training, which requires dense labeling on individual points or voxels in 3D scenes, thus intricate contextual perception is required to accomplish this element-wise recognition task. This characteristic makes semantic segmentation a promising foundation for further exploring scene-wise and object-wise recognition tasks, \ie, classification and detection.

\mypara{Dataset.} In our initial investigation into multi-dataset collaborative learning for 3D perception, we consider ScanNet~\cite{dai2017scannet}, S3DIS~\cite{armeni2016s3dis}, and Structured3D~\cite{zheng2020structured3d} as the datasets of interest, all of which include segmentation annotations. ScanNet and S3DIS represent the most commonly used real-world datasets in the realm of 3D perception, while Structured3D is a larger-scale synthetic RGB-D dataset that we specifically incorporated to establish an experimental context for addressing the domain gap between synthetic and real data, ultimately aiming to achieve mutual gains across datasets.
As illustrated in the left side of \tabref{tab:pilot_transfer}, although all three datasets represent indoor point cloud scenes, they exhibit distinct characteristics in terms of data scale, scene variety, and point cloud density. Our objective is to examine methods for overcoming the domain gap among these diverse datasets, facilitating collaborative learning across multiple sources and thereby taking an essential step towards large-scale representation learning for 3D perception.

\mypara{Evaluation.}
As a proof of concept, we consider \textit{joint training} by default, in which the model is jointly trained on all datasets under the supervised setting, and directly evaluated on all datasets without fine-tuning. In the final experiments, we will also consider two standard transfer learning settings: 1) \textit{supervised pre-training}, where the model supervised pre-trained during joint training is further fine-tuned on the target dataset; and 2) \textit{unsupervised pre-training}, where the model is unsupervised pre-trained on all datasets, and fine-tuned on each target dataset for evaluation.

\subsection{Pilot Study: Uncovering the Negative Transfer}
\label{sec:pilot}
As a pioneering effort, MSC~\cite{wu2023masked} involved unsupervised pre-training using a combination of two indoor datasets, ScanNet~\cite{dai2017scannet} and Arikitscene~\cite{dehghan2021arkitscenes}. However, even with the addition of three times more data, the performance improvement over the single-dataset pre-training baseline on ScanNet was relatively limited. To investigate the underlying causes of this limited performance gain, we take a step back and reassess this phenomenon by studying a straightforward supervised multi-dataset learning setup, \ie, the \textit{joint training} setting aforementioned in \secref{sec:setup}.

\textit{Negative transfer}~\cite{caruana1997multitask} refers to the phenomenon where learning from one dataset may negatively impact the performance on another dataset due to differences in data distribution. Despite restricting our focus to indoor scene point clouds, a significant negative transfer occurs during direct multi-dataset mixed segmentation training. As illustrated in \tabref{tab:pilot_transfer} (right side), we conduct training by pairwise merging the three datasets as well as a combination of all, and evaluate the model's performance on each related individual dataset. The experimental results reveal that direct merging training data gives rise to negative transfer between datasets, underscoring the challenges associated with attaining effective collaborative learning across multiple datasets in the 3D domain.
\begin{figure*}
    \centering
    \includegraphics[width=1\textwidth]{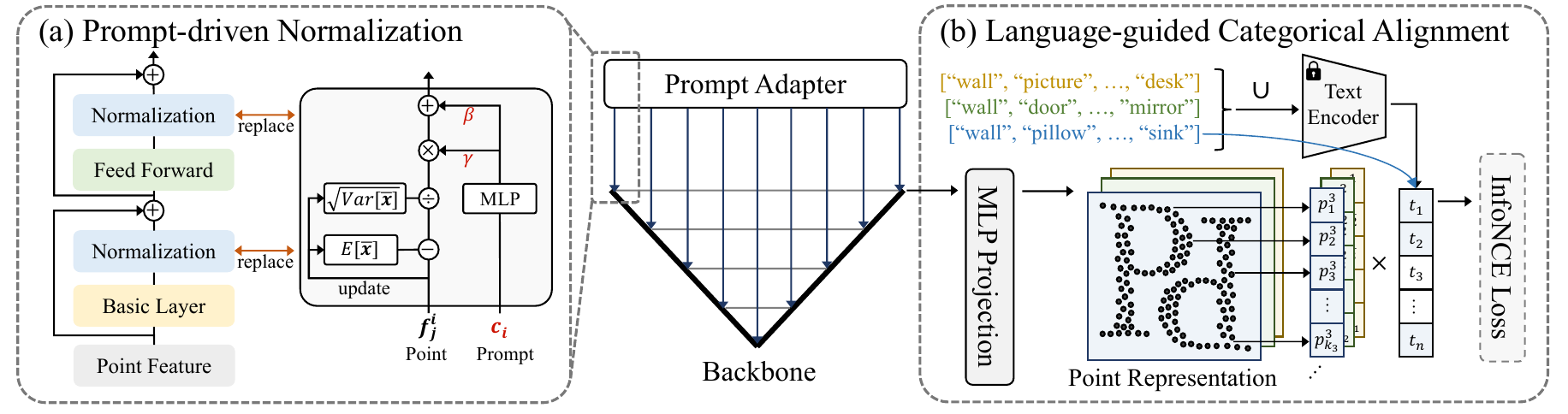}
    \caption{\textbf{Prompt adapter and categorical alignment.} (a) As a prompt adapter, \textit{Prompt-driven Normalization} adaptly encodes domain-specific prompts into the scale and shift vectors in normalization layers. This adaptation helps adapt the model to the specific dataset domain. (b) \textit{Language-guided Categorical Alignment}  aligns point representations to a unified category-language embedding, shared across all datasets and extracted by a pre-trained text encoder.}
    \vspace{-4mm}
    \label{fig:module}
\end{figure*}

\vspace{2mm}
\section{Point Prompt Training}
\vspace{2mm}
Due to the risk of negative transfer discussed in \secref{sec:pilot}, adapting a single model to diverse domains with distinct contexts still remains a significant challenge. Nevertheless, recent advances suggest that prompt tuning may be a viable approach for effectively adapting pre-trained models with large-scale datasets to downstream tasks. Inspired by this, we propose a different paradigm named Point Prompt Training (PPT) to mitigate negative transfer and enable multi-dataset training.

As shown in \figref{fig:module}, PPT has two essential components: (1) a prompt adapter, which adapts a single model to varying contexts of different datasets using a set of learnable domain-specific prompts, and (2) a categorical alignment process, which enables the model to be decently trained within multiple category spaces simultaneously with supervised learning. Details of them are presented as follows.

\vspace{1mm}
\subsection{Learning with Domain Prompting}
\vspace{1mm}

\mypara{Issues with prompt tuning.}
In the prompt tuning paradigm~\cite{liu2023pre}, a model pre-trained by a large-scale dataset is fine-tuned for specific tasks or datasets by incorporating additional information or context through prompts. These prompts facilitate the model's adaptation to new tasks with minimal parameter changes, often outperforming that with full fine-tuning~\cite{jia2022vpt, zhou2022coop, zhou2022cocoop} and laying the ground for a unified foundation model~\cite{brown2020gpt3}.

However, in 3D perception, the lack of a large-scale pre-trained model hinders the applications of prompt tuning. Furthermore, prompt tuning aims to address the domain gap between pre-training and fine-tuning datasets rather than improving the model's ability to fit multiple datasets simultaneously during either pre-training or fine-tuning. To tackle this issue, we introduce a novel method termed \textit{domain prompting}.
Instead of merely fine-tuning prompts on pre-trained models, we incorporate learnable prompt tokens as conditions for varying dataset contexts and (pre-)train the domain prompt with backbone cooperatively.

\mypara{Domain prompting.}
Specifically, for each interested dataset $\sD_i$, we generate a learnable $d$-dimensional vector as the domain-specific prompt. The collection of $n$ contexts is denoted as $\sC = \{\vc^i \in \mathbb{R}^d | i \in \mathbb{N}, 1 \leq i \leq n\}$. Then the multi-dataset training objective in \eqref{eq:multi-dataset-training} becomes:
\begin{align}
    \mathop{\operatorname{argmin}}_{\theta,\, \textcolor{red}{\sC}} \sum_{i=1}^n \frac{1}{|\sD_i|} \sum_{\left(\vx_j^i, \vy_j^i\right) \in \sD_i} \mathcal{L}\left(f\left(\vx_j^i, \textcolor{red}{\vc_i}; \theta\right), \vy_j^i\right) .
    \label{eq:conditional-multi-dataset-training}
\end{align}
These learnable domain prompts facilitate the discovery of distribution differences among datasets, enabling the backbone to surmount domain gaps encountered in multi-dataset training. As a result, the model focuses more on learning the representations that can be decently shared across datasets. This method fosters mutual benefits among distinct datasets and promotes a collaborative synergy between the backbone model and the prompts.
Similar to VPT~\cite{jia2022vpt}, we also observe that the shared prompt within each domain can achieve comparable or even better performance than the independent ones for different backbone blocks, and we put the discussion in the Appendix.
We believe this approach can benefit both supervised and unsupervised pre-training, as well as fine-tuning, by addressing the negative transfer that may exist within multiple datasets.

\mypara{Domain prompt adapter.} 
With the domain prompts that possess unique characteristics specific to individual datasets, enabling the model to effectively engage with domain-specific prompts becomes another challenge.
Previous research on visual prompt tuning has demonstrated that the adapters utilizing shared prompts to exert block-wise control over models are more effective than those that inject prompts at the input level~\cite{jia2022vpt}. Building on this insight, we investigate various designs for prompt adapters as outlined below and mark our main proposal with $*$. More specific illustrations and details regarding the alternative designs are available in our Appendix.
\begin{itemize}[leftmargin=5mm, itemsep=0mm, topsep=0mm, partopsep=-1mm]
    \item \textit{Direct Injection.} The domain-specific contextual cues of various datasets are encoded within their respective prompts. The incorporation of domain priors can be achieved by simply adding channel-aligned prompts to the intermediate feature maps with a linear projection.
    
    \item \textit{Cross Attention.} Drawing inspiration from DETR~\cite{detr}, we leverage a cross-attention-based domain prompt adapter as another alternative design for multi-dataset training. This scheme introduces a cross-attention block with a skip connection at the beginning of each encoder-decoder stage, injecting domain-specific information into the intermediate feature maps. Our design allows broad applicability to versatile 3D backbones without structural constraints while still preserving the advantages of the VPT technique.
    
    \item \textit{Prompt-driven Normalization$*$.} The objective of domain prompt adapter is to learn a shared representation that is robust and generalizable across various datasets, akin to how the style transfer methods\cite{dumoulin2016learned, ulyanov2017improved} retain the content essence while only transferring the contextual styles across images. Also, adapting the normalization layer to varying individual contexts is found beneficial for achieving better style transfer performance~\cite{huang2017arbitrary, peebles2022scalable}. With the analogy to style transfer, we introduce the context adapter of \textit{Prompt-driven Normalization} (PDNorm), a novel approach to tackle the transfer challenges associated with multi-dataset training illustrated in \figref{fig:module}\textcolor{red}{a}.
    Formally, with a given domain prompt $c$, PDNorm adaptively learns the $\gamma$ and $\beta$ values in normalization:
    \begin{align}
        \mathop{\operatorname{PDNorm}}(\vx, \textcolor{red}{\vc}) =\frac{\vx - \mathop{\operatorname{E}}[\bar{\vx}]}{\sqrt{\mathop{\operatorname{Var}}[\bar{\vx}] + \epsilon}} \cdot \gamma(\textcolor{red}{\vc}) + \beta(\textcolor{red}{\vc}),
    \label{eq:ada-norm}
    \end{align}
    where $\gamma(\vc)$ and $\beta(\vc)$ are linear projections, $\bar{\vx}$ for computing $\mathop{\operatorname{E}}[\bar{\vx}]$ and $\mathop{\operatorname{Var}}[\bar{\vx}]$ is contingent on the specific normalization employed by the backbone. It's important to note that $\mathop{\operatorname{E}}[\bar{\vx}]$ and $\mathop{\operatorname{Var}}[\bar{\vx}]$ are statisticized independently for each dataset involved. We substitute the original backbone's normalization layers with PdNorm layers. This approach promotes a more efficient yet effective alignment of feature distributions across datasets in the scenario of multi-dataset training.
\end{itemize}
\vspace{2mm}
\mypara{Zero-initialization and learning rate scaling. } Unlike prevalent prompt tuning methods that only adjust inserted prompts while retaining the pre-trained models, our proposed domain prompts are joint-trained with the backbone. Nevertheless, in our paradigm, the introduction of randomly initialized prompts may disrupt the representation learning of the rest of the model, resulting in unstable training with large loss values at early training stages. We conjecture that, during the initial stages of training, the model is acquiring general knowledge that can be applied across diverse domains. However, as training proceeds, the model gradually begins to generate domain-specific representations based on general representations. To address this issue, we employ zero-initialization~\cite{ioffe2015batch} and learning rate scaling~\cite{he2016deep}, ensuring stability during early training stages and yielding superior results. Specifically, we zero-initialize the $\gamma(\vc)$ and $\beta(\vc)$ parameters of PDNorm, and we start with a smaller base learning rate of prompt-related parameters to prioritize the backbone during the initial training stage. We also perform a similar design to our alternative prompt adapters for a fair comparison, and details are shown in the Appendix.

\subsection{Categorical Alignment}
In PPT, an additional critical issue that needs to be addressed is the inconsistency of the label space across different datasets with supervised learning. To tackle this problem, we have investigated various approaches to unify the categories for multi-dataset training as follows. Also, more details and discussions can be found in the Appendix.

\begin{itemize}[leftmargin=5mm, itemsep=0mm, topsep=-1mm, partopsep=-1mm]
    \item \textit{Decoupled.} One straightforward approach is to employ separate linear projection heads for each dataset. While this method is effective in handling inconsistencies, it introduces redundant parameters for decoding the same categories shared by different datasets. Besides, it overlooks the commonalities among the datasets and fails to account for their potential correlations.
    \item \textit{Unionized.} Another intuitive approach is to construct a shared linear segmentation head that projects the representation space into a unified label space encompassing all datasets while the loss computation remains separate and constrained to the distinct label spaces for each dataset. This method effectively resolves the inconsistency in point representations pertaining to the shared label space across datasets.
    \item \textit{Language-guided$*$.} 
    The aforementioned options treat each category independently and assume that they are uncorrelated. However, it is a natural fact that labels with close meanings should have similar representations~\cite{rozenberszki2022language}. Leveraging such prior information can further benefit the discovery of robust representations in our scenario. To this end, we propose language-guided categorical alignment, which aligns projected point representations with the category-language embeddings extracted by a pre-trained text encoder, such as CLIP~\cite{radford2021learning}. To achieve this goal, we employ the InfoNCE~\cite{oord2018representation} as alignment criteria and restrict negative samples to the specific dataset category space as shown in \figref{fig:module}\textcolor{red}{b}.
\end{itemize}

\section{Experiments}
In this section, we conduct extensive experiments to substantiate the efficacy of our proposed framework across multiple data sources with different evaluation settings. Specifically, in \secref{sec:ablation}, assess the effectiveness of different design choices via detailed ablation studies. After that, in \secref{sec:results}, we conduct system-level comparisons with existing methods. All experiments are conducted on compute nodes equipped with 8 NVIDIA A100 GPUs.

\begin{table*}[t!]
    \vspace{-3mm}
    \begin{minipage}{0.46\textwidth}
        \centering
        \tablestyle{0pt}{1.08}
        \begin{tabular}{x{34pt}| x{38pt} x{38pt} x{38pt} x{38pt}}\toprule
case &none &add &c.a. &\cellcolor[HTML]{efefef}p.n. \\\midrule
joint &\underline{68.9}  &70.9 &73.5 &\textbf{75.7} \\
f.t. &\underline{73.6} &73.8 &75.4 &\textbf{76.4} \\
\bottomrule
\end{tabular}
        \subcaption{\textbf{Prompt adapter.} Here \textit{c.a.} denotes \textit{cross attention}, and \textit{p.n.} denotes \textit{prompt-driven normalization}. \textit{p.n.} is a superior domain prompt adapter.}\label{subtab:adapter}
    \end{minipage}
    \hspace{8mm}
    \centering
    \begin{minipage}{0.48\textwidth}
        \hspace{3mm}
    \begin{minipage}{0.43\textwidth}
        \centering
        \tablestyle{0pt}{1.08}
        \begin{tabular}{x{33pt}| x{23pt} x{23pt}}\toprule
zero-init &w/o &\cellcolor[HTML]{efefef}w/ \\\midrule
joint &75.2  &\textbf{75.7} \\
f.t. &75.6 &\textbf{76.4} \\
\bottomrule
\end{tabular}
    \end{minipage}
    \hspace{-5mm}
    \begin{minipage}{0.46\textwidth}
        \centering
        \tablestyle{0pt}{1.08}
        \begin{tabular}{x{33pt}| x{23pt} x{23pt} x{23pt}}\toprule
lr scaler &1 &\cellcolor[HTML]{efefef}0.1 &0.01 \\\midrule
joint &75.4 &\textbf{75.7} &75.2 \\
f.t. &76.0 &\textbf{76.4} &75.8 \\
\bottomrule
\end{tabular}
    \end{minipage}
    \subcaption{\textbf{Zero-initialization \& Learning rate scaling.} Zero-init. benefits our training scheme, and 0.1 is a suitable lr scaler for domain prompting.}\label{subtab:zero}
    \end{minipage} \\
    \vspace{0mm}
    \begin{minipage}{0.46\textwidth}
        \centering
        \tablestyle{0pt}{1.08}
        \begin{tabular}{x{34pt}| x{38pt} x{38pt} x{38pt} x{38pt}}\toprule
location &initial &encoder &decoder &\cellcolor[HTML]{efefef}all \\\midrule
joint &68.7  &74.2 &73.2 &\textbf{75.7} \\
f.t. &73.9 &74.9 &74.7 &\textbf{76.4} \\
\bottomrule
\end{tabular}
        \subcaption{\textbf{Prompt location.} An in-depth prompt adapter that runs through the entire backbone is necessary.}\label{subtab:location}
    \end{minipage} 
    \hspace{8mm}
    \begin{minipage}{0.46\textwidth}
        \centering
        \tablestyle{0pt}{1.08}
        \begin{tabular}{x{34pt}| x{38pt} x{38pt} x{38pt} x{38pt}}\toprule
length &128 &\cellcolor[HTML]{efefef}256 &512 &1024 \\\midrule
joint &75.2  &\textbf{75.7} &\textbf{75.7} &75.5 \\
f.t. &75.9 &\textbf{76.4} &76.1 &76.2 \\
\bottomrule
\end{tabular}
        \subcaption{\textbf{Prompt length.} Domain prompt with 256 dimensions achieves a good balance.}\label{subtab:length}
    \end{minipage} \\
    \vspace{0mm}
    \begin{minipage}{0.46\textwidth}
        \centering
        \tablestyle{0pt}{1.08}
        \begin{tabular}{x{32pt}| x{40pt} x{38pt} x{38pt} x{38pt}}\toprule
case &decoupled &unionized &\cellcolor[HTML]{efefef}l.g. &l.g. w/ tpl. \\\midrule
joint &74.4  &75.3 &75.7 &\textbf{75.8} \\
f.t. &74.7 &75.8 &\textbf{76.4} &76.0 \\
\bottomrule
\end{tabular}
        \subcaption{\textbf{Categorical alignment.} \textit{l.g.} denotes \textit{languages-guided categorical alignment}, which achieves best f.t. results, and \textit{tpl.} stands for \textit{template}.}\label{subtab:alignment}
    \end{minipage}
    \hspace{8mm}
    \begin{minipage}{0.46\textwidth}
        \centering
        \tablestyle{0pt}{1.08}
        \begin{tabular}{x{34pt}| x{38pt} x{38pt} x{38pt} x{38pt}}\toprule
criteria &L2 &Disc.~\cite{de2017semantic} &TC~\cite{rozenberszki2022language} &\cellcolor[HTML]{efefef}InfoNCE \\\midrule
joint &12.8  &65.4 &70.2 &\textbf{75.7} \\
f.t. &72.1 & 73.5 &73.2 &\textbf{76.4} \\
\bottomrule
\end{tabular}
        \subcaption{\textbf{Language-guidance criteria.} \textit{TC} represents \textit{Text-supervised Contrastive} loss. While \textit{InfoNCE} loss makes \textit{l.g.} the best C.A..}\label{subtab:criteria}
    \end{minipage} \\
    \vspace{0mm}
    \begin{minipage}{0.46\textwidth}
        \centering
        \tablestyle{0pt}{1.08}
        \begin{tabular}{y{38pt}| x{37pt} x{37pt} x{37pt} x{37pt}}\toprule
ratio &\cellcolor[HTML]{efefef}4:2:1 &2:2:1 &2:1:1 &1:1:1 \\\midrule
ScanNet &75.7 &\textbf{75.8} &73.3 &74.6 \\
S3DIS &\textbf{72.2} &71.9 &71.2 &71.9 \\
Struct.3D &\textbf{75.8} &73.5 &72.7 &74.7 \\
\bottomrule
\end{tabular}
         \subcaption{\textbf{Sampling ratio.} The ratios below indicate the sampling rate of Struct.3D, ScanNet, and S3DIS, and 4:2:1 achieved the best result.}\label{subtab:data_ratio}
    \end{minipage}
    \hspace{8mm}
    \begin{minipage}{0.47\textwidth}
        \centering
        \tablestyle{0pt}{1.08}
        \begin{tabular}{y{38pt}| x{37pt} x{37pt} x{37pt} x{37pt}}\toprule
data &ScanNet &S3DIS &Struct.3D &\cellcolor[HTML]{efefef} all \\\midrule
    ScanNet &\underline{72.2}  &73.9 &\textbf{76.0} &75.7 \\
S3DIS &69.6 &\underline{65.4} &70.2 &\textbf{72.2} \\
Struct.3D &74.8 &75.0 &\underline{74.5} &\textbf{75.8} \\
\bottomrule
\end{tabular}
        \subcaption{\textbf{Joint training data.} Joint training results with different joint training data similar to \tabref{tab:pilot_transfer}. Datasets benefit each other in our PPT framework.}\label{subtab:data_co}
    \end{minipage}
    \vspace{-2mm}
    \caption{\textbf{Module ablation.} We adopt \textit{SparseUNet} and supervised multi-dataset \textit{joint training} to ablate our designs. We report both joint training and fine-tuning mIoU (\%) results on ScanNet 20-category semantic segmentation. All of our designs are enabled by default, and default settings are marked in \colorbox{gray}{gray}. The detailed setting for joint training and fine-tuning is reported in Appendix.}
    \label{tab:ablation}
    \vspace{-4mm}
\end{table*}

\begin{table*}[!t]
    \begin{minipage}{1\textwidth}
    \centering
        \vspace{-2mm}
        \tablestyle{10pt}{1.08}
        \begin{tabular}{lclllllll}\toprule
\multicolumn{2}{c}{Indoor Sem. Seg.} &\multicolumn{2}{c}{ScanNet~\cite{dai2017scannet}} &\multicolumn{2}{c}{ScanNet200~\cite{rozenberszki2022language}} &\multicolumn{2}{c}{S3DIS Area5~\cite{armeni2016s3dis}} \\\cmidrule(lr){3-4} \cmidrule(lr){5-6} \cmidrule(lr){7-8}
Methods &Params. &Val mIoU &Test mIoU &Val mIoU &Test mIoU &mIoU &mAcc \\\midrule
StratifiedFormer~\cite{lai2022stratified} &18.8M &74.3 &73.7 &- &- &72.0 &78.1 \\
PointNeXt~\cite{qian2022pointnext} &41.6M &71.5 &71.2 &- &- &70.5 &77.2 \\
PTv1~\cite{zhao2021point} &11.4M &70.6 &- &27.8 &- &70.4 &76.5 \\
PTv2~\cite{wu2022point} &12.8M &75.4 &75.2 &30.2 &- &71.6 &77.9 \\ \midrule
SparseUNet~\cite{choy20194d} &39.2M &\underline{72.2} &\underline{73.6} &\underline{25.0} &\underline{25.3} &\underline{65.4} &\underline{71.7} \\
+ PC~\cite{xie2020pointcontrast} &39.2M &74.1$\,_\text{\textcolor{darkgray}{(+1.9)}}$ &- &26.2$\,_\text{\textcolor{darkgray}{(+1.2)}}$ &- &70.3$\,_\text{\textcolor{darkgray}{(+4.9)}}$ &76.9$\,_\text{\textcolor{darkgray}{(+5.2)}}$ \\
+ CSC~\cite{hou2021exploring} &39.2M &73.8$\,_\text{\textcolor{darkgray}{(+1.6)}}$ &- &26.4$\,_\text{\textcolor{darkgray}{(+1.4)}}$ &24.9$\,_\text{\textcolor{darkgray}{(-0.4)}}$ &72.2$\,_\text{\textcolor{darkgray}{(+6.8)}}$ &- \\
+ MSC~\cite{wu2023masked} &39.2M &75.5$\,_\text{\textcolor{darkgray}{(+3.3)}}$ &- &28.8$\,_\text{\textcolor{darkgray}{(+3.8)}}$ &- &70.1$\,_\text{\textcolor{darkgray}{(+4.7)}}$ &77.2$\,_\text{\textcolor{darkgray}{(+5.5)}}$ \\
\rowcolor{gray}+ PPT Unsup. (f.t.) &41.0M &75.8$\,_\text{\textcolor{darkgreen}{(+3.6)}}$ &- &30.4$\,_\text{\textcolor{darkgreen}{(+5.4)}}$ &- &71.9$\,_\text{\textcolor{darkgreen}{(+6.5)}}$ &\textbf{78.3}$\,_\text{\textcolor{darkgreen}{(+6.6)}}$ \\
\rowcolor{gray}+ PPT Sup. (joint) &41.0M &75.7$\,_\text{\textcolor{darkgreen}{(+3.5)}}$ &\textbf{76.6}$\,_\text{\textcolor{darkgreen}{(+3.0)}}$ &- &- &72.2$\,_\text{\textcolor{darkgreen}{(+6.8)}}$ &78.0$\,_\text{\textcolor{darkgreen}{(+6.3)}}$ \\
\rowcolor{gray}+ PPT Sup. (f.t.) &41.0M &\textbf{76.4}$\,_\text{\textcolor{darkgreen}{(+4.2)}}$ &- &\textbf{31.9}$\,_\text{\textcolor{darkgreen}{(+6.9)}}$ &\textbf{33.2}$\,_\text{\textcolor{darkgreen}{(+7.9)}}$ &\textbf{72.7}$\,_\text{\textcolor{darkgreen}{(+7.3)}}$ &78.2$\,_\text{\textcolor{darkgreen}{(+6.5)}}$ \\\midrule
PTv3~\cite{pointcept2023} &46.2M &77.5 &77.9 &35.2 &37.8 &73.4 &77.7 \\
\rowcolor{gray}+ PPT Sup. (f.t.) &46.3M &\textbf{78.6} $\,_\text{\textcolor{darkgreen}{(+1.1)}}$ &\textbf{78.3} 
$\,_\text{\textcolor{darkgreen}{(+0.4)}}$ &\textbf{36.0} 
$\,_\text{\textcolor{darkgreen}{(+0.8)}}$ &\textbf{39.3} 
$\,_\text{\textcolor{darkgreen}{(+1.5)}}$ &\textbf{74.7} 
$\,_\text{\textcolor{darkgreen}{(+1.3)}}$ &\textbf{79.6} 
$\,_\text{\textcolor{darkgreen}{(+1.9)}}$ \\
\bottomrule
\end{tabular}
        \vspace{-2mm}
        \caption{\textbf{Indoor semantic segmentation results.} Our method builds on SparseUNet~\cite{choy20194d} and PTv3~\cite{pointcept2023}, and is evaluated on ScanNet, ScanNet200, and S3DIS benchmarks. The framework is universal, and we report on three settings: unsupervised pre-training integrated with MSC~\cite{wu2023masked}, supervised joint training, and supervised pre-training. Besides comparing with previous pre-training methods~\cite{xie2020pointcontrast,hou2021exploring,wu2023masked}, we also conduct system-level comparisons against previous SOTAs~\cite{lai2022stratified,qian2022pointnext,zhao2021point,wu2022point}, and our method shows consistently better results across benchmarks even with one single share-weighted model.}\label{tab:sem_seg}
        \vspace{1mm}
    \end{minipage} \\
    \begin{minipage}{1\textwidth}
    \centering
        \tablestyle{10pt}{1.08}
        \begin{tabular}{lclllllll}\toprule
\multicolumn{2}{c}{Outdoor Sem. Seg.} &\multicolumn{2}{c}{SemanticKITTI~\cite{behley2019semantickitti}} &\multicolumn{2}{c}{nuScenes~\cite{caesar2020nuscenes}} &\multicolumn{2}{c}{Waymo~\cite{sun2020waymo}} \\\cmidrule(lr){3-4} \cmidrule(lr){5-6} \cmidrule(lr){7-8}
Methods &Params. &Val mIoU &Test mIoU &Val mIoU &Test mIoU &Val mIoU &Val mAcc \\\midrule
SPVNAS~\cite{tang2020spvnas} &10.8M &64.7 &66.4 &- &77.4 &- &- \\
Cylinder3D~\cite{zhu2021cylindrical} &26.1M &64.3 &67.8 &76.1  &77.2 &- &- \\
SphereFormer~\cite{lai2023spherical} &32.3M &67.8 &74.8 &78.4 &81.9 &69.9 &- \\ \midrule
SparseUNet~\cite{choy20194d} &39.2M &\underline{63.8} &- &\underline{73.3} &- &\underline{65.9} &\underline{76.6} \\
\rowcolor{gray}+ PPT Sup. (joint) &41.0M &70.9$\,_\text{\textcolor{darkgreen}{(+7.1)}}$ &- &78.5$\,_\text{\textcolor{darkgreen}{(+5.2)}}$ &- &70.0$\,_\text{\textcolor{darkgreen}{(+4.1)}}$ &\textbf{79.1}$\,_\text{\textcolor{darkgreen}{(+2.5)}}$ \\
\rowcolor{gray}+ PPT Sup. (f.t.) &41.0M &\textbf{71.4}$\,_\text{\textcolor{darkgreen}{(+7.6)}}$ &- &\textbf{78.6}$\,_\text{\textcolor{darkgreen}{(+5.3)}}$ &- &\textbf{70.4}$\,_\text{\textcolor{darkgreen}{(+4.5)}}$ &78.9$\,_\text{\textcolor{darkgreen}{(+2.3)}}$ \\ \midrule
PTv3~\cite{pointcept2023} &46.2M &70.8 &74.2 &80.4 &82.7  &71.3 &80.5 \\
\rowcolor{gray}+ PPT Sup. (f.t.) &46.3M &\cellcolor[HTML]{efefef}\textbf{72.3} 
$\,_\text{\textcolor{darkgreen}{(+1.5)}}$ &\cellcolor[HTML]{efefef}\textbf{75.5} 
$\,_\text{\textcolor{darkgreen}{(+1.3)}}$ &\cellcolor[HTML]{efefef}\textbf{81.2} 
$\,_\text{\textcolor{darkgreen}{(+0.8)}}$ &\cellcolor[HTML]{efefef}\textbf{83.0} 
$\,_\text{\textcolor{darkgreen}{(+0.3)}}$ &\cellcolor[HTML]{efefef}\textbf{72.1} 
$\,_\text{\textcolor{darkgreen}{(+0.8)}}$ &\cellcolor[HTML]{efefef}\textbf{81.3} 
$\,_\text{\textcolor{darkgreen}{(+0.8)}}$ \\
\bottomrule
\end{tabular}
        \vspace{-2mm}
        \caption{\textbf{Outdoor semantic segmentation results.} We also examine the efficacy of PPT in an outdoor context using SparseUNet~\cite{choy20194d} and PTv3~\cite{pointcept2023}. Our evaluation encompasses SemanticKITTI, nuScenes, and Waymo semantic segmentation benchmarks. We report on two main settings: supervised joint training and supervised pre-training. We conduct comprehensive comparisons against previous SOTAs~\cite{tang2020spvnas,zhu2021cylindrical,wu2022point,lai2023spherical}, and our method shows multiple superior results across benchmarks.}\label{tab:sem_seg_outdoor}
        \vspace{-5mm}
    \end{minipage}
\end{table*}

\subsection{Ablation Study}
\label{sec:ablation}
In this part, we ablate different design choices of PPT from the perspective of module design and data engineering. We employ supervised joint training with \textit{SparseUNet}, train it on ScanNet, S3DIS, and Structured3D, and evaluate it on ScanNet 20-category semantic segmentation. For evaluation, we consider both direct evaluation (joint training) and fine-tuning (see details in \secref{sec:setup}). More details of the setting are available in the Appendix.

\mypara{Prompt adapter.}
In \tabref{subtab:adapter}, we show results with different designs of the domain prompt adapter. Compared with the vanilla baseline (none) without a prompt adapter, all designs show effectiveness in learning good representations from multiple datasets. Moreover, compared with simpler designs like direct injection (add) and cross attention (c.a.), our novel design prompt-driven normalization (p.n.) achieves significantly stronger results, verifying its effectiveness.

\mypara{Zero-initialization and learning rate scaling.}
In \tabref{subtab:zero}, we verify the effect of zero initialization and learning rate scaling. Overall, it shows that zero initialization, a technique often adopted for adapting pre-trained models, could also benefit training from scratch. Besides, scaling the learning rate for domain prompting to a relatively smaller value (0.1) than the backbone also helps training.

\mypara{Prompt location.}
In \tabref{subtab:location}, we study the influence of injecting the prompt adapter to different stages of the backbone. Empirically, the benefit of the prompt adapter becomes higher if it is added to relatively deeper stages. Our intuition is that features in earlier stages are more related to low-level attributes, which could be easier shared across datasets. And, deeper features are more related to high-level semantics, where negative effect of the domain gap occurs and a domain adapter is needed.

\mypara{Prompt length.}
In \tabref{subtab:length}, we ablate the feature-level length (dimension) of the prompt adapter. A larger dimension of the adapter often allows space for higher information capability, but our experiments show that the adapter is quite memory-efficient. The results with different feature dimensions do not differ much, and a small dimension of 256 is already sufficient.

\mypara{Categorical alignment.}
In \tabref{subtab:alignment}, we show results with different methods for aligning the label space of different training datasets. Compared with learning separate segmentation heads for each dataset, obtaining a unionized head allows better alignment of the supervision from different datasets. Further, language guidance takes the relationship between class names into account, resolving possible conflicts, and results in a further performance boost. Besides that, we also tried a simple prompt engineering technique that augments class names to a sentence (\eg, ``A point of [class].''), which does not show effectiveness in this case.

\mypara{Language-guidance criteria.}
In \tabref{subtab:criteria}, we ablate the loss function for aligning with category-specific language embeddings extracted from a pre-trained text encoder. Simple L2 loss, which does not consider negative examples, could result in mode collapse. Compared with other specialized criteria, \eg, text-supervised contrastive loss proposed in~\cite{rozenberszki2022language}, our method suits well with the most commonly used InfoNCE loss, highlighting its universality.

\mypara{Sampling ratio.}
In \tabref{subtab:data_ratio}, we show the results with different sampling ratios across datasets, and experiments show that overall our method is relatively robust to this ratio. It is important to note that, in contrast to downstream tasks where the sampling ratio can significantly impact the final performance, our focus is on representation learning. Therefore, the effect of the sampling ratio may be negligible if the model is sufficiently trained on each dataset for an adequate duration~\cite{he2019rethinking}.

\mypara{Joint training data.}
In \tabref{subtab:data_co}, we show the results with different joint training data (see attributes of datasets in \tabref{tab:pilot_transfer}). Note that though they differ in data source, sparsity, complexity, and scale, our final framework allows consistent benefit from different data sources regardless of large domain gaps.

\subsection{Results Comparision}
\label{sec:results}

\begin{table*}[!t]
    \begin{minipage}{1\textwidth}
    \centering
        \vspace{-2mm}
        \tablestyle{12pt}{1.08}
        \begin{tabular}{lclllllll}\toprule
\multicolumn{2}{c}{Indoor Ins. Seg.} &\multicolumn{3}{c}{ScanNet Val~\cite{dai2017scannet}} &\multicolumn{3}{c}{ScanNet200 Val~\cite{rozenberszki2022language}} \\\cmidrule(lr){3-5} \cmidrule(lr){6-8}
PointGroup~\cite{jiang2020pointgroup} &Params. &mAP@25 &mAP@50 &mAP\ \ \  &mAP@25 &mAP@50 &mAP\ \ \  \\\midrule
SparseUNet~\cite{choy20194d} &39.2M &\underline{72.8} &\underline{56.9} &\underline{36.0} &\underline{32.2} &\underline{24.5} &\underline{15.8} \\
+ PC~\cite{xie2020pointcontrast} &39.2M & - &58.0$\,_\text{\textcolor{darkgray}{(+1.1)}}$ &- &- &24.9$\,_\text{\textcolor{darkgray}{(+0.4)}}$ &- \\
+ CSC~\cite{hou2021exploring} &39.2M & - &59.4$\,_\text{\textcolor{darkgray}{(+2.5)}}$ &- &- &25.2$\,_\text{\textcolor{darkgray}{(+0.7)}}$ &- \\
+ LGround~\cite{rozenberszki2022language} &39.2M & - &- &- &- &26.1$\,_\text{\textcolor{darkgray}{(+1.6)}}$ &- \\
+ MSC~\cite{wu2023masked} &39.2M &74.7$\,_\text{\textcolor{darkgray}{(+1.9)}}$ &59.6$\,_\text{\textcolor{darkgray}{(+2.7)}}$ &39.3$\,_\text{\textcolor{darkgray}{(+3.3)}}$ &34.3$\,_\text{\textcolor{darkgray}{(+2.1)}}$ &26.8$\,_\text{\textcolor{darkgray}{(+2.3)}}$ &17.3$\,_\text{\textcolor{darkgray}{(+1.5)}}$\\
\rowcolor{gray}+ PPT (f.t.) &41.0M &\textbf{76.9}$\,_\text{\textcolor{darkgreen}{(+4.1)}}$ &\textbf{62.0}$\,_\text{\textcolor{darkgreen}{(+3.1)}}$ &\textbf{40.7}$\,_\text{\textcolor{darkgreen}{(+4.7)}}$ &\textbf{36.8}$\,_\text{\textcolor{darkgreen}{(+4.6)}}$ &\textbf{29.4}$\,_\text{\textcolor{darkgreen}{(+4.9)}}$ &\textbf{19.4}$\,_\text{\textcolor{darkgreen}{(+3.6)}}$ \\ \midrule
PTv3~\cite{pointcept2023} &46.2M &77.5 &61.7 &40.9 &40.1 &33.2 &23.1 \\
\rowcolor{gray}+ PPT (f.t.) &46.3M &\textbf{78.9}$\,_\text{\textcolor{darkgreen}{(+1.4)}}$ &\textbf{63.5}$\,_\text{\textcolor{darkgreen}{(+1.8)}}$ &\textbf{42.1}$\,_\text{\textcolor{darkgreen}{(+1.2)}}$ &\textbf{40.8}$\,_\text{\textcolor{darkgreen}{(+0.7)}}$ &\textbf{34.1}$\,_\text{\textcolor{darkgreen}{(+0.9)}}$ &\textbf{24.0}$\,_\text{\textcolor{darkgreen}{(+0.9)}}$ \\
\bottomrule
\end{tabular}
        \vspace{-2mm}
        \caption{\textbf{Indoor instance segmentation results.} We conduct PPT supervised pre-training on SparseUNet~\cite{choy20194d} as described in \tabref{tab:sem_seg} and further fine-tuning on ScanNet and ScanNet200 instance segmentation driven by PointGroup~\cite{jiang2020pointgroup}. We compare mAP@25, mAP@50, and mAP results with previous pre-training methods, and our method shows significant superior results across benchmarks}\label{tab:ins_seg}
    \end{minipage}
    \nextfloat
    \vspace{-1mm}
    \begin{minipage}{0.48\textwidth}
        \vspace{0.5mm}
        \centering
        \tablestyle{2pt}{1.08}
        \begin{tabular}{x{30pt}| x{30pt} x{42pt} x{42pt} x{42pt}}\toprule
Pct. &SC &CSC~\cite{hou2021exploring} &MSC~\cite{wu2023masked} &\cellcolor[HTML]{efefef}PPT \\\midrule
1\% &26.0  &28.9$\,_\text{\textcolor{darkgray}{(+2.9)}}$ &29.2$\,_\text{\textcolor{darkgray}{(+3.2)}}$ &\textbf{31.3}$\,_\text{\textcolor{darkgreen}{(+5.3)}}$ \\
5\% &47.8  &49.8$\,_\text{\textcolor{darkgray}{(+2.0)}}$ &50.7$\,_\text{\textcolor{darkgray}{(+2.9)}}$ &\textbf{52.2}$\,_\text{\textcolor{darkgreen}{(+4.4)}}$ \\
10\% &56.7  &59.4$\,_\text{\textcolor{darkgray}{(+2.7)}}$ &61.0$\,_\text{\textcolor{darkgray}{(+4.3)}}$ &\textbf{62.8}$\,_\text{\textcolor{darkgreen}{(+6.1)}}$ \\
20\% &62.9  &64.6$\,_\text{\textcolor{darkgray}{(+1.7)}}$ &64.9$\,_\text{\textcolor{darkgray}{(+2.0)}}$ &\textbf{66.4}$\,_\text{\textcolor{darkgreen}{(+3.5)}}$ \\
100\% &72.2  &73.8$\,_\text{\textcolor{darkgray}{(+1.6)}}$ &75.3$\,_\text{\textcolor{darkgray}{(+3.1)}}$ &\textbf{75.8}$\,_\text{\textcolor{darkgreen}{(+3.6)}}$ \\
\bottomrule
\end{tabular}
         \subcaption{\textbf{Limited reconstructions.} \textit{Pct.} denotes the percentage of scene reconstruction that could be used for training.}\label{subtab:lr}
    \end{minipage}
    \hspace{6mm}
    \begin{minipage}{0.48\textwidth}
        \centering
        \vspace{0.5mm}
        \tablestyle{2pt}{1.08}
        \begin{tabular}{x{30pt}| x{30pt} x{42pt} x{42pt} x{42pt}}\toprule
Pts. &SC &CSC~\cite{hou2021exploring} &MSC~\cite{wu2023masked} &\cellcolor[HTML]{efefef}PPT \\\midrule
20 &41.9  &55.5$\,_\text{\textcolor{darkgray}{(+13.6)}}$ &60.1$\,_\text{\textcolor{darkgray}{(+18.2)}}$ &\textbf{60.6}$\,_\text{\textcolor{darkgreen}{(+18.7)}}$ \\
50 &53.9  &60.5$\,_\text{\textcolor{darkgray}{(+6.6)} }$ &66.8$\,_\text{\textcolor{darkgray}{(+12.9)}}$ &\textbf{67.5}$\,_\text{\textcolor{darkgreen}{(+13.6)}}$ \\
100 &62.2  &65.9$\,_\text{\textcolor{darkgray}{(+3.7)} }$ &69.7$\,_\text{\textcolor{darkgray}{(+7.5)} }$ &\textbf{70.8}$\,_\text{\textcolor{darkgreen}{(+8.6)} }$ \\
200 &65.5  &68.2$\,_\text{\textcolor{darkgray}{(+2.7)} }$ &70.7$\,_\text{\textcolor{darkgray}{(+5.2)} }$ &\textbf{72.2}$\,_\text{\textcolor{darkgreen}{(+6.7)} }$ \\
Full &72.2  &73.8$\,_\text{\textcolor{darkgray}{(+1.6) }}$ &75.3$\,_\text{\textcolor{darkgray}{(+3.1)} }$ &\textbf{75.8}$\,_\text{\textcolor{darkgreen}{(+3.6)} }$ \\
\bottomrule
\end{tabular}
        \subcaption{\textbf{Limited annotations.} \textit{Pts.} denotes the number of points per scene that are annotated for training.}\label{subtab:la}
    \end{minipage}
    \vspace{-2mm}
    \caption{\textbf{Data efficient results.} We follow the ScanNet Data Efficient benchmark~\cite{hou2021exploring} and compare the validation results of the PPT unsupervised setting with previous pre-training methods. All methods are trained with SparseUNet, and \textit{SC} denotes train from scratch.}\label{tab:data_efficient}
    \vspace{-5mm}
\end{table*}

\mypara{Indoor semantic segmentation results.}
In \tabref{tab:sem_seg}, we present the main results of different variants of our method on multiple standard semantic segmentation benchmarks, and compare with previous state-of-the-art methods at both system-level and module-level.
Following the common practice of pre-training methods~\cite{xie2020pointcontrast,hou2021exploring,wu2023masked}, our method is built on both convolution-based architecture SparseUNet~\cite{choy20194d} and transformer-based architecture PTv3~\cite{pointcept2023}.
Under the unsupervised setting, our framework could smoothly integrate MSC~\cite{wu2023masked} and enable it to benefit from joint training on multiple datasets, \eg, improving on ScanNet200 Val split by 1.6 points, and on S3DIS Area5 mIoU by 1.8 points. More importantly, the results also surpass all previous SOTAs, verifying the effectiveness and potential of large-scale unsupervised pre-training for 3D scene understanding. When further considering the supervised joint training setting, and fine-tuning upon it, our method further sees consistent performance gains across tasks and secures position as a new SOTA.

\mypara{Outdoor semantic segmentation results.}
In \tabref{tab:sem_seg_outdoor}, we expand our methodology to outdoor scenarios by presenting additional results of our approach on multiple outdoor semantic segmentation benchmarks. We systematically compare these results with those of previously established SOTA methods. Our method is still based on SparseUNet \cite{choy20194d}, a classic framework within the outdoor perception community, and PTv3~\cite{pointcept2023}, which is the latest SOTA backbone for outdoor perception. Under the supervised joint training paradigm, our method showcases significant enhancements across all tasks when contrasted with scratch results, even with a single shared-weight model. For instance, on the SemanticKITTI Validation split, our approach elevates by 7.1 points, underscoring the potential of all-data learning in the realm of 3D understanding. Through subsequent fine-tuning on each dataset, PPT consistently demonstrates superiority over the latest literature. For instance, it outperforms SphereFormer~\cite{lai2023spherical} by 5.0 points in terms of mIoU on the SemanticKITTI validation set.

\mypara{Indoor instance segmentation results.}
In \tabref{tab:ins_seg}, we conduct fine-tuning experiments on instance segmentation using SparseUNet~\cite{choy20194d} and PTv3~\cite{pointcept2023} as the backbone, powered by PointGroup~\cite{jiang2020pointgroup}. The fine-tuning outcomes are reported on both the ScanNet~\cite{dai2017scannet} and ScanNet200~\cite{rozenberszki2022language} instance segmentation benchmarks. Our findings consistently reveal the superior performance of our approach compared to the prior state-of-the-art method, MSC~\cite{wu2023masked}. To be specific, PPT outperforms MSC by 2.4 points in terms of mAP@50 on the ScanNet validation split, and by 2.6 points on the ScanNet200 validation split. This underscores the effectiveness of the point representation learned by PPT in enhancing instance segmentation performance.

\mypara{Data efficient benchmark.}
In \tabref{tab:data_efficient}, we report results for the ScanNet Data Efficient benchmark~\cite{hou2021exploring}, where scene reconstruction or annotation percentages are limited. Our method, integrating MSC~\cite{wu2023masked}, is compared with prior pre-training methods and consistently outperforms them under data efficient settings.

\vspace{-2mm}
\section{Conclusion and Discussion}
\vspace{-1mm}

This paper introduces PPT, an effort toward large-scale 3D representation learning with a novel 3D multi-dataset synergistic training setting. We identify the negative transfer issue and present a unified framework that addresses this problem with the proposed Prompt-driven Normalization and Language-guided Categorical Alignment, delivering consistent and significant performance gains.

We discuss \textit{limitations and broader impacts} as follows:
\begin{itemize}[leftmargin=5mm, itemsep=0mm, topsep=-1mm, partopsep=-1mm]
\item \textit{Module design.} As a preliminary work in 3D multi-dataset pre-training, this paper first verifies the effectiveness of this setting and opens doors for large-scale 3D representation learning. Yet current explorations are still restricted to a limited scope and the designs could be sub-optimal, thus further study on more advanced techniques is necessary. For example, one could verify the effectiveness of this framework when combined with more advanced unsupervised pre-training methods and explore more effective prompting techniques.
\item \textit{Data domain.} Our study demonstrates the potential benefit of simultaneously utilizing both synthetic and real point cloud data. It would be exciting to see this ability extended to more specific scenarios in different domains, \eg, jointly learning from both indoor and outdoor scenes.
\item \textit{Multi-task training.} Our current formulation only considers one pre-training task. Upon that, as it has shown the ability to achieve superior results across datasets with a single model, a promising direction is to enable multi-task training for 3D scene understanding with a unified framework.
\end{itemize}

\vspace{-2mm}
\section*{Acknowledgements}
\vspace{-2mm}
This work is supported in part by the National Natural Science Foundation of China (No. 622014840), Alibaba Innovative Research Fund, HKU Startup Fund, and HKU Seed Fund for Basic Research.

\appendix
\section*{Appendix}
For a thorough understanding of our Point Prompt Training (PPT), we have compiled a detailed Appendix. The table of contents below offers a quick overview and will guide to specific sections of interest.

\hypersetup{linkbordercolor=black,linkcolor=black}

\setlength{\cftbeforesecskip}{0.5em}
\cftsetindents{section}{0em}{1.8em}
\cftsetindents{subsection}{1em}{2.5em}
\cftsetindents{subsubsection}{2em}{2.5em}

\etoctoccontentsline{part}{Appendix}
\localtableofcontents
\hypersetup{linkbordercolor=red,linkcolor=red}

\begin{figure*}[!th]
    \centering
    \includegraphics[width=1\textwidth]{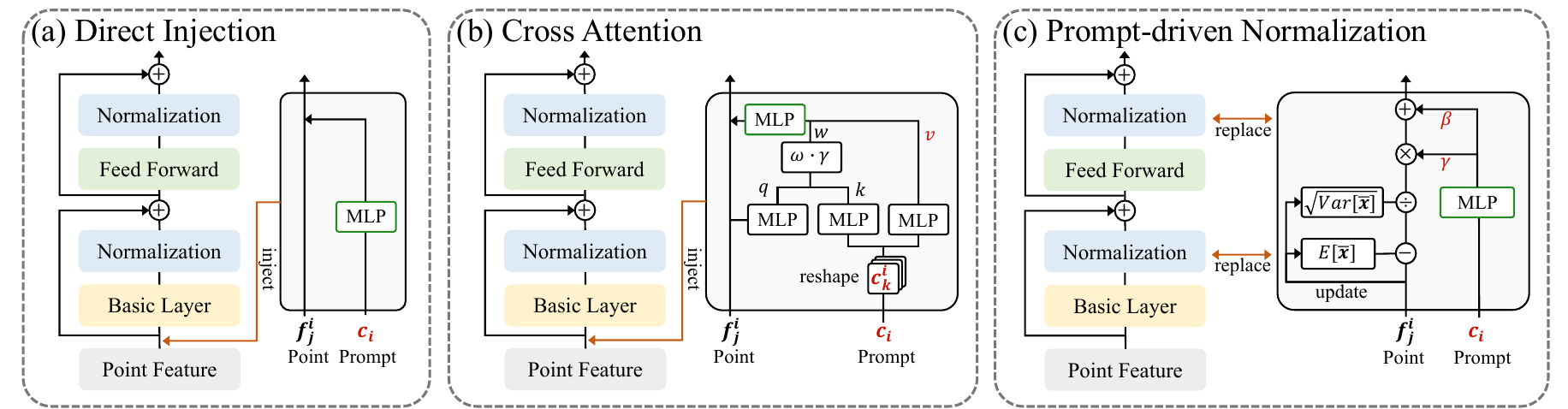}
    \vspace{-6mm}
    \caption{\textbf{Domain prompt adapters.}}
    \label{fig:adapters}
    \vspace{2mm}
    \includegraphics[width=1\textwidth]{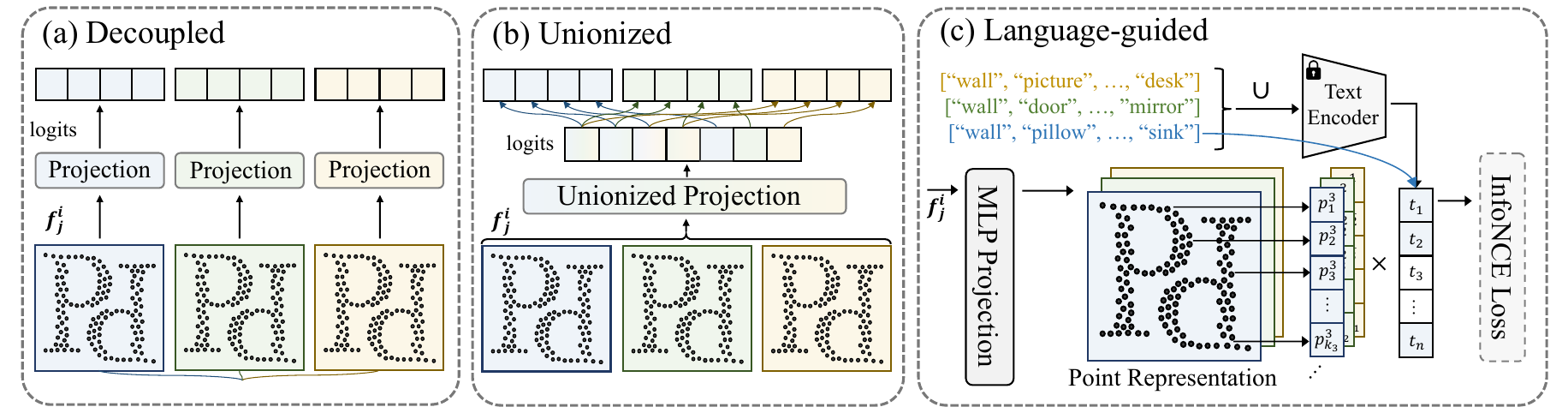}
    \vspace{-6mm}
    \caption{\textbf{Categorical alignment approaches.}}
    \label{fig:alignments}
\end{figure*}

\phantomsection
\section{Related Work}

\addcontentsline{toc}{subsubsection}{(1)\ \ \ \ 3D Scene Understanding}
\mypara{3D scene understanding.}
Deep learning techniques for understanding 3D scenes using neural networks can be broadly classified into three categories based on their approach to handling point clouds: projection-based, voxel-based, and point-based methods.
Projection-based approaches involve projecting 3D points onto multiple image planes and utilizing 2D CNN-based backbones for feature extraction~\cite{su15mvcnn,li2016vehicle,chen2017multi,lang2019pointpillars}.
In contrast, voxel-based methods convert point clouds into regular voxel representations to facilitate 3D convolutions
~\cite{maturana2015voxnet,song2017semantic}.
The efficiency of these methods is further enhanced through the use of sparse convolution techniques~\cite{graham20183d,choy20194d}.
Unlike the previous two, point-based methods operate directly on point clouds~\cite{qi2017pointnet,qi2017pointnet++, zhao2019pointweb,thomas2019kpconv} and have recently begun incorporating transformer-based architectures~\cite{guo2021pct,zhao2021point,wu2022point}.
Following previous pre-training literatures~\cite{xie2020pointcontrast,hou2021exploring,wu2023masked}, we train on the voxel-based SparseUNet~\cite{choy20194d}, which is more efficient and allows large-scale training.

\addcontentsline{toc}{subsubsection}{(2)\ \ \ \ 3D Representation Learning}
\mypara{3D representation learning.}
Deep neural networks are notoriously data-hungry, and scaling up the pre-training data has become a promising path to learning robust and transferrable representations.
Unlike in 2D vision, where large-scale curated datasets are readily available~\cite{deng2009imagenet,asano2021pass}, data collection and annotation in 3D vision is much more costly, and the scale of point cloud datasets are quite limited~\cite{dai2017scannet,armeni2016s3dis}.
Regarding 3D representation learning, previous works commonly pre-train on a single dataset~\cite{wang2019deep,hassani2019unsupervised,sauder2019self,sanghi2020info3d,xie2020pointcontrast,hou2021exploring}, which limits the potential to benefit from the scaling law~\cite{kaplan2020scaling}.
As the first attempt towards scaling up the pre-training data, a recent work~\cite{wu2023masked} first explored unsupervised pre-training on merged data (ScanNet~\cite{dai2017scannet} and ArkitScenes~\cite{dehghan2021arkitscenes}).
However, as the distributions of 3D datasets vary much, naively merging them could be sub-optimal, which is studied in this work.

\addcontentsline{toc}{subsubsection}{(3)\ \ \ \ Towards Large-scale Pre-training}
\mypara{Towards large-scale pre-training.}
In order to scale up pre-training and learn better representations, two popular topics in 2D vision is to exploit uncurated data in the wild~\cite{thomee2016yfcc100m,sun2017jft,tian2021divide,goyal2021self}, and to better utilize the data in hand~\cite{zhao2020object,xie2021pixpro,liu2020self,xie2021unsupervised,wen2022self}.
Yet the former is not applicable to 3D data, and the latter has been well-studied in previous works~\cite{xie2020pointcontrast,hou2021exploring,wu2023masked}.
The topic of joint learning across multiple datasets has also been explored in some works related to 2D scene understanding~\cite{kim2022learning,wang2022cross,zhou2022simple,yao2023detclipv2,vandenhende2021multi} and 3D object detection~\cite{zhang2023uni3d}, but while they focus on direct evaluation on the target dataset (similar to domain generalization~\cite{ranftl2022midas,wang2022generalizing,chen2023scaledet, cha2021swad}). Our work targets more on generalized representation learning in both supervised and unsupervised settings.
Moreover, the high variation between 3D datasets, and the sparse and heavily long-tailed nature, also add to the difficulty of 3D joint training.

\addcontentsline{toc}{subsubsection}{(4)\ \ \ \ Prompt Learning}
\mypara{Prompt learning.} 
In an effort to improve the generalizability of pre-trained models on downstream tasks, prompting was originally proposed in natural language processing~\cite{liu2023pre}.
The prompt templates could be heuristic designed~\cite{brown2020gpt3,schick2021pet,han2022ptr}, automatically generated~\cite{shin2020autoprompt,gao2021making}, or learned as task-specific parameters~\cite{lester2021power,liu2021gpt,gu2022ppt,hu2022knowledgeable,cui2022prototypical}. We rephrase the latter one as prompt learning.
In 2D vision, prompt learning has become a popular parameter-efficient technique to adapt pre-trained models to specific \textit{downstream tasks}~\cite{jia2022vpt,zhou2022coop,zang2022unified,ju2022prompting,bahng2022visual,ge2022domain}.
Our work, instead, tackles \textit{pre-training} directly. Prompt learning is regarded as a dataset-specific adapter to allow the model to resolve the domain shift between pre-training datasets separately, and learn the optimal overall representation.

\section{Alternative Designs}
In this section, we provide a comprehensive overview and visual demonstration of the implementation details for proposed alternative designs for Point Prompt Training (PPT).

\subsection{Domain Prompt Adapter}
To address the challenges of adapting the model to different dataset domains, we introduce domain prompt adapters along with zero-initialization techniques. \figref{fig:adapters} serves as a visual guide, showcasing the implementation of each domain prompt adapter discussed in the main paper. Notably, the zero-initialized layers are highlighted with a \textcolor{darkgreen}{green} box.

\phantomsection
\addcontentsline{toc}{subsubsection}{(1)\ \ \ \ Direct Indiction}
\mypara{Direct Indiction.} \figref{fig:adapters}\textcolor{red}{a} showcases the process of Direct Injection. This approach inserts a direct injection adapter at the beginning of each basic block. The domain prompt is added to the point embedding after undergoing a zero-initialized linear projection within each direct injection.

\addcontentsline{toc}{subsubsection}{(2)\ \ \ \ Cross Attention}
\mypara{Cross Attention.} As shown in \figref{fig:adapters}\textcolor{red}{b}, the cross-attention adapter can be seen as an extension of the direct injection adapter. The domain prompts $c_i$ splits into $k$ independent prompt embeddings of identical shape, serving as the reference for cross-attention with each point. Attention operations~\cite{wu2022point} occur between query vectors from each point and key value vectors from the prompt embeddings. The output, post-projection by a zero-initialized linear layer, is added to the point embedding.

\addcontentsline{toc}{subsubsection}{(3)\ \ \ \ Prompt-driven Normalization}
\mypara{Prompt-driven Normalization.} \figref{fig:adapters}\textcolor{red}{c} illustrates the Prompt-driven Normalization (PDNorm) approach. In this case, each normalization layer is replaced with PDNorm, which enables the adaptation of the backbone to the specific domain context. PDNorm projects the domain prompt onto the scale-shift vector using a zero-initialized linear layer, and these domain-aware vectors are subsequently applied to the normalized feature embedding.

\subsection{Categorical Alignment}
To address the issue of inconsistency within the category space during supervised multi-dataset ergiergistic training, various categorical alignment strategies are explored in the main paper. \figref{fig:alignments} provides a detailed illustration of these categorical alignment methods.

\begin{table*}[!th]
    \begin{minipage}{1\textwidth}
        \centering
        \tablestyle{2.5pt}{1.08}
        \begin{tabular}{lc|cccccccccccccccccccccccccccccccccccc}
    \toprule
    Datasets &\#C & \rotatebox{90}{wall} &
    \rotatebox{90}{floor} & \rotatebox{90}{cabinet} & \rotatebox{90}{bed} & \rotatebox{90}{chair} & \rotatebox{90}{sofa} & \rotatebox{90}{table} & \rotatebox{90}{door} & \rotatebox{90}{window} & \rotatebox{90}{bookshelf} & \rotatebox{90}{bookcase} & \rotatebox{90}{picture} & \rotatebox{90}{counter} & \rotatebox{90}{desk} & \rotatebox{90}{shelves} & \rotatebox{90}{curtain} & \rotatebox{90}{dresser} & \rotatebox{90}{pillow} & \rotatebox{90}{mirror} & \rotatebox{90}{ceiling} & \rotatebox{90}{refrigerator} & \rotatebox{90}{television} & \rotatebox{90}{shower curtain} & \rotatebox{90}{nightstand} & \rotatebox{90}{toilet} & \rotatebox{90}{sink} & \rotatebox{90}{lamp} & \rotatebox{90}{bathtub} & \rotatebox{90}{garbagebin} & \rotatebox{90}{board} & \rotatebox{90}{beam} & \rotatebox{90}{column} & \rotatebox{90}{clutter} & \rotatebox{90}{otherstructure} & \rotatebox{90}{otherfurniture} & \rotatebox{90}{otherprop} \\\midrule
    
    ScanNet &20 &\checkmark &\checkmark &\checkmark &\checkmark &\checkmark &\checkmark &\checkmark &\checkmark &\checkmark &\checkmark & &\checkmark &\checkmark &\checkmark & &\checkmark & & & & &\checkmark & &\checkmark & &\checkmark &\checkmark & &\checkmark & & & & && &\checkmark & \\
    
    S3DIS & 13 &\checkmark &\checkmark & & &\checkmark &\checkmark &\checkmark &\checkmark &\checkmark & &\checkmark & & & & & & & & &\checkmark & & & & & & & & & &\checkmark &\checkmark &\checkmark &\checkmark & & & \\
    
    Struct.3D & 25 &\checkmark &\checkmark &\checkmark &\checkmark &\checkmark &\checkmark &\checkmark &\checkmark &\checkmark & & &\checkmark & &\checkmark &\checkmark &\checkmark &\checkmark &\checkmark &\checkmark &\checkmark &\checkmark &\checkmark & &\checkmark & &\checkmark &\checkmark & & & & & & &\checkmark & \checkmark & \checkmark \\
\bottomrule
\end{tabular}
        \vspace{-2mm}
        \caption{\textbf{Categorical settings.} }
        \label{tab:appendix_categories}
    \end{minipage}
    \begin{minipage}{1\textwidth}
        \centering
        \tablestyle{9pt}{1.08}
        \begin{tabular}{x{60pt} x{70pt} x{60pt} x{70pt} x{60pt} x{45pt}}\toprule
\multicolumn{2}{c}{\textbf{Pre-training} (joint)} &\multicolumn{2}{c}{\textbf{Fine-tuning} (ScanNet)} &\multicolumn{2}{c}{\textbf{Fine-tuning} (S3DIS)} \\\cmidrule(lr){1-2} \cmidrule(lr){3-4} \cmidrule(lr){5-6}
\textbf{Config} &\textbf{Value} &\textbf{Config} &\textbf{Value} &\textbf{Config} &\textbf{Value} \\
\cmidrule(lr){1-1} \cmidrule(lr){2-2} \cmidrule(lr){3-3} \cmidrule(lr){4-4} \cmidrule(lr){5-5} \cmidrule(lr){6-6}
optimizer &SGD &optimizer &SGD &optimizer &SGD \\
scheduler &cosine decay &scheduler &cosine decay &scheduler &poly \\
learning rate &0.5 &learning rate &0.5 &learning rate &0.1 \\
weight decay &1e-4 &weight decay &1e-4 &weight decay &1e-4 \\
momentum &0.8 &momentum &0.9 &momentum &0.9 \\
batch size &24 &batch size &12 &batch size &12 \\
datasets &ScanNet~(2) &datasets &ScanNet &datasets &S3DIS \\
&S3DIS~(1) & &- & &- \\
&Struct.3D~(4) & &- & &- \\
warmup iters &6k &warmup epochs &40 &warmup epochs &0 \\
iters &120k &epochs &800 &epochs &3000 \\
\bottomrule
\end{tabular}
        \vspace{-2mm}
        \caption{\textbf{Training settings.} }
        \label{tab:training_settings}
    \end{minipage} \\
    \vspace{-6mm}
\end{table*}

\phantomsection
\addcontentsline{toc}{subsubsection}{(1)\ \ \ \ Decoupled}
\mypara{Decoupled.} \figref{fig:alignments}\textcolor{red}{a} shows the decoupled approach for categorical alignment. In this method, a separate prediction head is employed for each dataset. After the shared backbone extracts the point embeddings, they are fed into the prediction head specific to the corresponding dataset's domain. Loss calculation is performed within the category space corresponding to each domain.

\addcontentsline{toc}{subsubsection}{(2)\ \ \ \ Unionzied}
\mypara{Unionzied.} \figref{fig:alignments}\textcolor{red}{b} presents the unified method of categorical alignment. Unlike the decoupled strategy, point embeddings are not split based on their respective domains. Instead, they pass through a unified prediction head that projects the point representations into the unified category space. The logit value of each category is predicted within this space. However, we still restrict each point prediction space to its corresponding domain's category space during loss calculation.

\addcontentsline{toc}{subsubsection}{(3)\ \ \ \ Language-guided}
\mypara{Language-guided.} \figref{fig:alignments}\textcolor{red}{c} demonstrates the language-guided approach. Here, we leverage a CLIP~\cite{radford2021learning} pre-trained text encoder to extract the text embedding of each category. The alignment process involves aligning each point representation with the text embedding of its category. This alignment is facilitated by utilizing InfoNCE~\cite{oord2018representation} loss as the alignment criterion. Specifically, we calculate the similarity between the point representation and the text embedding. The resulting similarity matrix is multiplied by a logit scaler (100)~\cite{radford2021learning} to determine the logit value of each category, and cross-entropy loss is computed accordingly.

\section{Additional Experiments}
\subsection{Experimental Settings}
\phantomsection
\addcontentsline{toc}{subsubsection}{(1)\ \ \ \ Data}
\mypara{Data.} 
We conduct PPT joint (pre)-training on three datasets: ScanNet v2~\cite{dai2017scannet}, S3DIS~\cite{armeni2016s3dis}, and Structured3D~\cite{zheng2020structured3d}. The ScanNet v2 dataset consists of 1,613 scene scans reconstructed from RGB-D frames. It is partitioned into 1,201 scenes for training, 312 scenes for validation, and 100 scenes for benchmark testing. Point clouds in this dataset are sampled from the vertices of reconstructed meshes, and each sampled point is assigned a semantic label from a set of 20 categories. The S3DIS dataset comprises 271 rooms from six areas in three distinct buildings. Model performance evaluation is typically done using results from Area 5 and 6-fold cross-validation (result available in \tabref{tab:s3dis-6fold}). Unlike ScanNet v2, points in the S3DIS dataset are densely sampled on the surfaces of the meshes and annotated into 13 categories. Structured3D is a synthetic photo-realistic dataset containing 3.5K house designs created by professional designers. It is annotated with the same set of 40 categories as the NYU Depth V2~\cite{silberman2012indoor} dataset. The dataset is divided into 3,000 scenes for training, 250 scenes for validation, and 250 scenes for testing. We further split the 3,500 scenes into approximately 20,000 rooms and project the panoramic image of each room into a 3D point cloud for training. Following the approach in Swin3D~\cite{yang2023swin3d}, the frequency of occurrence of the 40 categories is counted. Categories with frequencies less than 0.001 are filtered out, and end up with a reduced set of 25 categories for perception. Similar to Swin3D, we include the categories table of the three datasets in \tabref{tab:appendix_categories} to provide a clear reference to the category relation across the three datasets.

\addcontentsline{toc}{subsubsection}{(2)\ \ \ \ Training}
\mypara{Training.} The default joint (pre-)training and fine-tuning setting is in \tabref{tab:training_settings}. During joint training, we follow a sampling strategy where the batched point cloud for each iteration was sampled from a single dataset. The sampling ratio is determined based on the best performance necessary iteration number for each dataset. This approach ensures that each dataset contributes to the training process in proportion to its optimal performance. Consequently, the total number of training iterations is equal to the sum of the best performance necessary iteration numbers for all the datasets involved as mentioned above. Furthermore, we observe that using a larger batch size leads to more stable performance during training. Our fine-tuning follows the practice of supervised SparseUNet training setting from \textit{Pointcept}~\cite{pointcept2023}.

\addcontentsline{toc}{subsubsection}{(3)\ \ \ \ Backbone}
\mypara{Backbone.} We validate the effectiveness of our Point Prompt Training by leveraging SparseUNet~\cite{choy20194d}, optimized by \textit{Pointcept}~\cite{pointcept2023} with the \textit{SpConv}~\cite{spconv2022} library. The utilization of SparseUNet was chosen due to its notable advantages in terms of speed and memory efficiency. The specific configuration of the backbone is outlined in ~\tabref{tab:backbone_settings}, with our primary results based on the widely employed SpUNet-S, featuring 39 million parameters. Additionally, we explore the impact of employing a larger-scale backbone with 412 million parameters, denoted as SpUNet-L. The analysis of PPT's properties with the larger-scale backbone is discussed in \secref{sec:additional_ablation}.

\begin{table}[!t]
    \begin{minipage}{0.48\textwidth}
        \centering
        \tablestyle{5pt}{1.08}
        \begin{tabular}{ccc}\toprule
\textbf{Config} &\multicolumn{2}{c}{\textbf{Value}} \\
\cmidrule(lr){1-1} \cmidrule(lr){2-3}
\rowcolor{gray} name &SpUNet-S~(default) &SpUNet-L \\
patch embed depth &1 &1 \\
patch embed channels &32 &96 \\
patch embed kernel size &5 &5 \\
encode depths &[2, 3, 4, 6] &[6, 6, 12, 6] \\
encode channels &[32, 64, 128, 256] &[96, 192, 384, 768] \\
encode kernel size &3 &3 \\
decode depths &[2, 2, 2, 2] &[2, 2, 2, 2] \\
decode channels &[256, 128, 64, 64] &[768, 384, 192, 192] \\
decode kernel size &3 &3 \\
pooling stride &[2, 2, 2, 2] &[2, 2, 2, 2] \\
params &39M &412M \\
\bottomrule
\end{tabular}

        \vspace{-2mm}
        \caption{\textbf{Backbone settings.} }
        \label{tab:backbone_settings}
    \end{minipage}
    \vspace{-4mm}
\end{table}

\subsection{Additional Pilot Study}
\mypara{Naive joint-training with varied sampling ratios.}
In the pilot study, which is conducted in the main paper, we perform training experiments by naively pairwise merging ScanNet, S3DIS, and Structure3D datasets, as well as training on a combination of all datasets. Subsequently, we evaluate the model's performance on each individual dataset. The determination of the sampling ratio is based on the necessary iteration number for achieving the best performance on each dataset. Consequently, we select a sampling ratio of 4:2:1 for Structure3D, ScanNet, and S3DIS accordingly.

Concerns naturally arise regarding the potential impact of a larger sampling rate for the Structured3D point cloud. It is possible that this could lead to the model bias toward the more frequently witnessed domain, exacerbating performance degradation in other datasets rather than improving naively joint training. To investigate this further, we conduct an additional pilot study, exploring different sampling rates during naively joint training. 

\tabref{tab:appendix_sampling_ratios} provides an illustration of two representative sampling ratios: 4:2:1 and 1:1:1. The experimental results indicate that although increasing the sampling rate of ScanNet and S3DIS data with the balanced sampling ratio 1:1:1 slightly alleviated the performance degradation, the negative transfer effect remained significant in our vanilla setting. These findings further underscore the challenges associated with achieving effective collaborative learning across multiple datasets in the 3D domain.

\begin{table*}[t]
    \hspace{10mm}
    \begin{minipage}{0.43\textwidth}
        \centering
        \tablestyle{2pt}{1.08}
        \begin{tabular}{y{38pt}| x{37pt} x{37pt} x{37pt} x{37pt}}\toprule
data &ScanNet &S3DIS &Struct.3D &\cellcolor[HTML]{efefef} all \\\midrule
    ScanNet &\underline{72.2}  &69.5 &67.2 &69.7 \\
S3DIS &64.7 &\underline{65.4} &63.6 &63.5 \\
Struct.3D &73.9 &73.7 &\underline{74.5} &72.4 \\
\bottomrule
\end{tabular}

        \subcaption{Sampling Ratio 1:1:1}\label{subtab:data_ratio_111}
    \end{minipage}
    \hspace{0mm}
    \begin{minipage}{0.43\textwidth}
        \centering
        \tablestyle{2pt}{1.08}
        \begin{tabular}{y{38pt}| x{37pt} x{37pt} x{37pt} x{37pt}}\toprule
data &ScanNet &S3DIS &Struct.3D &\cellcolor[HTML]{efefef} all \\\midrule
    ScanNet &\underline{72.2}  &71.8 &65.9 &68.9 \\
S3DIS &64.1 &\underline{65.4} &62.8 &63.3 \\
Struct.3D &73.7 &74.2 &\underline{74.5} &72.9 \\
\bottomrule
\end{tabular}
        \subcaption{Sampling Ratio 4:2:1}\label{subtab:data_ratio_421}
    \end{minipage} \\
    \vspace{-3mm}
    \caption{\textbf{Naive joint-training with varied sampling ratios.}} 
    \label{tab:appendix_sampling_ratios}
    \vspace{2mm}
    \begin{minipage}{1\textwidth}
        \centering
        \tablestyle{13.5pt}{1.08}
        \begin{tabular}{c|cccccc|cc}\toprule
split &Area1 &Area2 &Area3 &Area4 &Area5 &Area6 &\cellcolor[HTML]{efefef}PPT &Scratch \\\midrule
mIoU &83.01 &65.39 &87.09 &74.13 &72.73 &86.42 &\textbf{78.13} &\underline{65.4} \\
mAcc &90.25 &75.58 &91.83 &84.01 &78.22 &92.47 &\textbf{85.39} &- \\
allAcc &93.48 &88.34 &94.56 &90.84 &91.45 &94.45 &\textbf{92.19} &- \\
\bottomrule
\end{tabular}
        \vspace{-3mm}
        \caption{\textbf{S3DIS semantic segmentation 6-fold cross-validation results.}}
        \label{tab:s3dis-6fold}
        \vspace{2mm}
    \end{minipage} \\
    \begin{minipage}{1\textwidth}
    \centering
        \tablestyle{8.5pt}{1.08}
        \begin{tabular}{lclllllll}\toprule
\multicolumn{2}{c}{} &\multicolumn{2}{c}{ScanNet~\cite{dai2017scannet}} &\multicolumn{2}{c}{ScanNet200~\cite{rozenberszki2022language}} &\multicolumn{2}{c}{S3DIS Area5~\cite{armeni2016s3dis}} \\\cmidrule(lr){3-4} \cmidrule(lr){5-6} \cmidrule(lr){7-8}
Methods &Params. &Val mIoU &Test mIoU &Val mIoU &Test mIoU &mIoU &mAcc \\\midrule
SparseUNet~\cite{choy20194d} &39.2M &\underline{72.2} &\underline{73.6} &\underline{25.0} &\underline{25.3} &\underline{65.4} &\underline{71.7} \\
\rowcolor{gray}+ PPT Sup. (joint) &41.0M &75.4$\,_{\textcolor{darkgray}{\pm\text{0.46}}}$ &- &- &- &71.9$\,_{\textcolor{darkgray}{\pm\text{0.32}}}$ &77.5$\,_{\textcolor{darkgray}{\pm\text{0.38}}}$ \\
\rowcolor{gray}+ PPT Sup. (f.t.) &41.0M &76.2$\,_{\textcolor{darkgray}{\pm\text{0.18}}}$ &- &31.7$\,_{\textcolor{darkgray}{\pm\text{0.22}}}$&- &72.4$\,_{\textcolor{darkgray}{\pm\text{0.21}}}$ &77.9$\,_{\textcolor{darkgray}{\pm\text{0.30}}}$ \\
\bottomrule
\end{tabular}
        \vspace{-3mm}
        \caption{\textbf{Error bar-supplemented results.}}\label{tab:mean_std}
        \vspace{2mm}
    \end{minipage} \\
    \nextfloat
    \begin{minipage}{0.37\textwidth}
        \centering
        \tablestyle{8pt}{1.08}
        \begin{tabular}{c|cccc}\toprule
backbone &S &L &S &L \\
PPT &- &- &\checkmark &\checkmark \\\midrule
results &73.4 &72.9 &75.7 &\textbf{75.8} \\
\bottomrule
\end{tabular}
        \subcaption{Backbone Up-scaling}\label{subtab:model_scale_up}
    \end{minipage}
    \hspace{0mm}
    \centering
    \begin{minipage}{0.22\textwidth}
        \centering
        \tablestyle{8pt}{1.08}
        \begin{tabular}{c|cc}\toprule
backbone &S &S \\
shared &- &\checkmark \\\midrule
results &75.3 & \textbf{75.7} \\
\bottomrule
\end{tabular}

        \subcaption{Shared Domain Prompt}\label{subtab:shared_prompt}
    \end{minipage}
    \hspace{0mm}
    \centering
    \begin{minipage}{0.26\textwidth}
        \centering
        \tablestyle{8pt}{1.08}
        \begin{tabular}{c|cc}\toprule
backbone &S &S \\
head &Linear &LCA \\\midrule
results &73.4 &\textbf{74.2} \\
\bottomrule
\end{tabular}

        \subcaption{LCA as Prediction Head}\label{subtab:lca_head}
    \end{minipage}
    \vspace{-3mm}
    \caption{\textbf{Additional ablation.}}
    \vspace{-4mm}
\end{table*}

\begin{figure*}[t]
    \centering
    \includegraphics[width=0.90\textwidth]{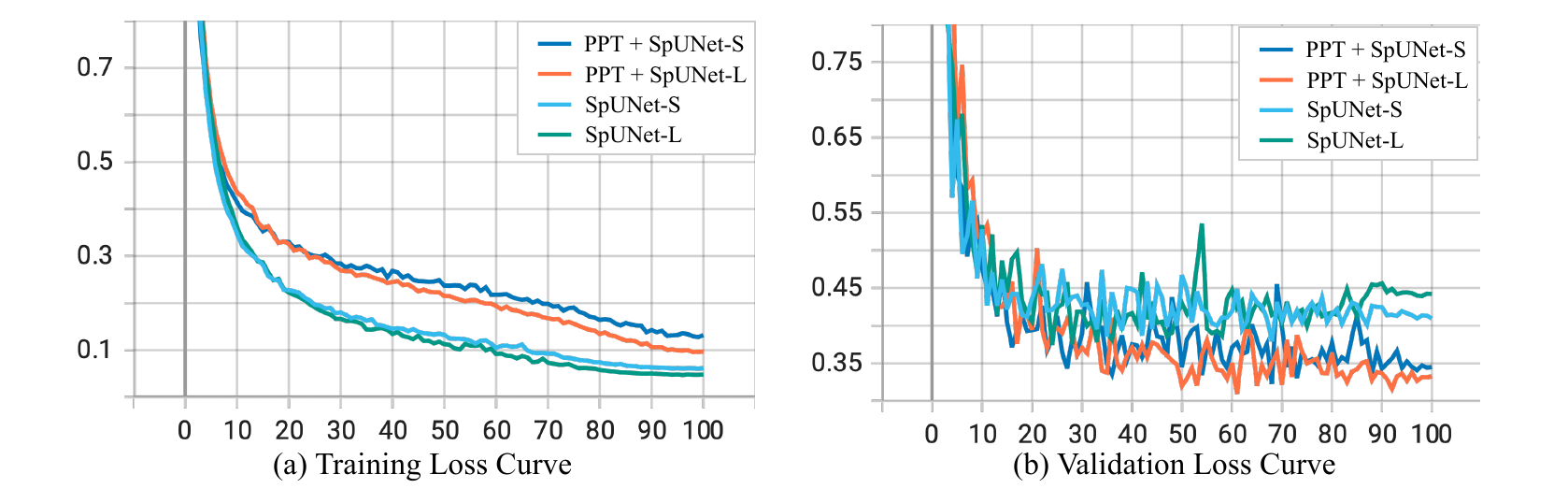}
    \vspace{-3mm}
    \caption{\textbf{Loss curve.}} \label{fig:loss_curve}
\end{figure*}

\subsection{Additional Results}
\phantomsection
\addcontentsline{toc}{subsubsection}{(1)\ \ \ \ S3DIS 6-fold Semantic Segmentation}
\mypara{S3DIS 6-fold semantic segmentation.} \tabref{tab:s3dis-6fold} presents the results of our 6-fold cross-validation semantic segmentation experiment on the S3DIS dataset. For each fold, we withhold one area of S3DIS and perform PPT joint training using the remaining data along with the ScanNet and Structured3D datasets. We then evaluate and report the model's performance on the withheld area data. The average of these results represents the 6-fold cross-validation results. Notably, Point Prompt Training achieves a significant improvement in SparseUNet performance on this benchmark, with a notable 12.7\% increase, establishing a new SOTA result.

\addcontentsline{toc}{subsubsection}{(2)\ \ \ \ Error Bar-supplemented Results}
\mypara{Error bar-supplemented results.} As a supplement to the main paper, we present the full semantic segmentation results in the main paper in \tabref{tab:mean_std}, in which we supplement the error bar derived from five independent runs. The mean-std result of the ScanNet test mIoU is not available since multiple submissions are not allowed.

\subsection{Additional Ablation Study}
\label{sec:additional_ablation}

\phantomsection
\addcontentsline{toc}{subsubsection}{(1)\ \ \ \ Backbone Up-scaling}
\mypara{Backbone up-scaling.}
\tabref{subtab:model_scale_up} presents our investigation into the impact of scaling up the backbone using multi-dataset Point Prompt Training (PPT). As a baseline, we evaluate the performance of SpUNet-S and SpUNet-L trained solely on the ScanNet dataset. Our observations indicate that, in this setup, increasing the model capacity results in significant overfitting. However, when PPT is introduced with a larger-scale data source, the issue of overfitting is mitigated, and a larger-scale backbone yields improved model performance.

To provide a visual representation of these findings, \figref{fig:loss_curve} illustrates the loss curves for the training and validation splits of the four experiments. The entire training period was evenly divided into 100 epochs, and the average loss on the training and validation splits was calculated at the end of each epoch to generate the curves. It is noteworthy that SpUNet-L with PPT exhibits a more favorable loss curve compared to SpUNet-S with PPT, while the opposite trend is observed in the absence of PPT.

However, it is important to consider that expanding the depth and dimension of convolution-based models results in a significant increase in parameters. As a result, transformer-based methods are better suited for exploring model capacity expansion. Nevertheless, it is worth noting that transformer-based methods currently have limitations in terms of speed and memory consumption. As part of future work, optimizing the efficiency of transformer-based backbones after scaling up remains a topic worth investigating. \looseness=-1

\begin{table}[!t]
    \begin{minipage}{0.48\textwidth}
    \centering
        \tablestyle{10pt}{1.08}
        \begin{tabular}{lccc}\toprule
Methods &Year &Val &Test \\\midrule
\scratch PointNet++~\cite{qi2017pointnet++} & 2017 & 53.5 & 55.7 \\
\scratch 3DMV~\cite{dai20183dmv} & 2018 & -& 48.4\\
\scratch PointCNN~\cite{li2018pointcnn} & 2018 & -& 45.8\\
\scratch SparseConvNet~\cite{graham20183d} & 2018 & 69.3 & 72.5 \\
\scratch PanopticFusion~\cite{narita2019panopticfusion} & 2019 & -& 52.9\\

\scratch PointConv~\cite{wu2019pointconv} & 2019 & 61.0 & 66.6\\
\scratch JointPointBased~\cite{chiang2019unified}& 2019 & 69.2 & 63.4\\
\scratch KPConv~\cite{thomas2019kpconv} & 2019 & 69.2 & 68.6 \\
\scratch PointASNL~\cite{yan2020pointasnl}& 2020 & 63.5 & 66.6\\
\scratch SegGCN~\cite{lei2020seggcn}& 2020 & -& 58.9\\
\scratch RandLA-Net~\cite{hu2020randla} & 2020 & - & 64.5 \\
\scratch JSENet~\cite{hu2020jsenet} & 2020 & - & 69.9 \\
\scratch FusionNet~\cite{zhang2020deep} & 2020 & - & 68.8 \\
\scratch PTv1~\cite{zhao2021point} &2021 &70.6 &- \\
\scratch FastPointTransformer~\cite{park2022fast} &2022 &72.4 &- \\
\scratch SratifiedTranformer~\cite{lai2022stratified} &2022 &74.3 &73.7 \\
\scratch PointNeXt~\cite{qian2022pointnext} &2022 &71.5 &71.2 \\
\scratch PTv2~\cite{wu2022point} &2022 &75.4 &74.2 \\
\scratch LargeKernel3D~\cite{chen2023largekernel3d} & 2023 &73.5 &73.9 \\
\scratch PointMetaBase~\cite{lin2023pointmetabase} & 2023 &72.8 &71.4 \\
\scratch PointConvFormer~\cite{wu2023pointconvformer} & 2023 &74.5 &74.9 \\
\scratch OctFormer~\cite{Wang2023OctFormer} &2023 &75.7 &76.6 \\
\scratch Swin3D~\cite{yang2023swin3d} &2023 &75.5 &- \\
\pretrain\ + Supervised~\cite{yang2023swin3d} &2023 &76.7 &77.9 \\
\scratch MinkUNet~\cite{choy20194d} &2019 &72.2 &73.6 \\
\pretrain\ + PC~\cite{xie2020pointcontrast} &2020 &74.1 &- \\
\pretrain\ + CSC~\cite{hou2021exploring} &2021 &73.8 &- \\
\pretrain\ + MSC~\cite{wu2023masked} &2024 &75.5 &- \\
\pretrain\ + GC~\cite{wang2024gc} &2024 &75.7 &- \\
\cellcolor[HTML]{efefef}\pretrain\ + PPT~(Ours) &\cellcolor[HTML]{efefef}2024 &\cellcolor[HTML]{efefef}76.4 &\cellcolor[HTML]{efefef}76.6 \\
\scratch OA-CNNs~\cite{peng2024oacnns} &2024 &76.1 &75.6 \\
\scratch PTv3~\cite{pointcept2023} &2024 &77.5 &77.9 \\
\cellcolor[HTML]{efefef}\pretrain\ + PPT (Ours) &\cellcolor[HTML]{efefef}2024 &\cellcolor[HTML]{efefef}\textbf{78.6} &\cellcolor[HTML]{efefef}\textbf{79.4} \\
\bottomrule
\end{tabular}
        \vspace{-2mm}
        \caption{\textbf{ScanNet V2 semantic segmentation.} }\label{tab:appendix_scannet}
        \vspace{-6mm}
    \end{minipage}
\end{table}

\addcontentsline{toc}{subsubsection}{(2)\ \ \ \ Shared Domain Prompt}
\mypara{Shared domain prompt.}
In \tabref{subtab:shared_prompt}, validating the effectiveness of globally shared domain prompts in comparison to independent ones across different backbone blocks.  Similar to the conclusion in VPT~\cite{jia2022vpt}, we observe that employing block-wise independent domain prompts resulted in a decline in performance. We attribute this to the complexity introduced by having separate domain prompts for each block, leading to overfitting. This aligns with the observations from our ablation study in the main paper, where scaling up prompt dimensions had a similar degradation.

\begin{table}[!t]
    \begin{minipage}{0.48\textwidth}
    \centering
        \tablestyle{10pt}{1.08}
        \begin{tabular}{lccc}\toprule
Methods &Year &Area5 &6-fold \\\midrule
\scratch PointNet~\cite{qi2017pointnet} &2017 &41.1 &47.6\\
\scratch SegCloud~\cite{tchapmi2017segcloud} &2017 &48.9 &-\\
\scratch TanConv~\cite{tatarchenko2018tangent} &2018 &52.6 &-\\
\scratch PointCNN~\cite{li2018pointcnn} &2018 &57.3 &65.4\\
\scratch ParamConv~\cite{wang2018deep} &2018 &58.3 &-\\
\scratch PointWeb~\cite{zhao2019pointweb} &2019 &60.3 &66.7\\
\scratch HPEIN~\cite{jiang2019hierarchical} &2019 &61.9 &-\\
\scratch KPConv~\cite{thomas2019kpconv} &2019 &67.1 &70.6\\
\scratch GACNet~\cite{wang2019graph} &2019 &62.9 &-\\
\scratch PAT~\cite{yang2019modeling} &2019 &60.1 &-\\
\scratch SPGraph~\cite{landrieu2018large} &2018 &58.0 &62.1\\
\scratch SegGCN~\cite{lei2020seggcn} &2020 &63.6 &-\\
\scratch PAConv~\cite{xu2021paconv} &2021 &66.6 &-\\
\scratch PTv1~\cite{zhao2021point} &2021 &70.4 &65.4 \\
\scratch StratifiedTransformer~\cite{lai2022stratified} &2022 &72.0 &- \\
\scratch PointNeXt~\cite{qian2022pointnext} &2022 &70.5 &74.9 \\
\scratch PTv2~\cite{wu2022point} &2022 &71.6 &73.5 \\
\scratch PointMetaBase~\cite{lin2023pointmetabase} & 2023 &72.0 &77.0 \\
\scratch Swin3D~\cite{yang2023swin3d} &2023 &72.5 &76.9 \\
\pretrain\ + Supervised~\cite{yang2023swin3d} &2023 &74.5 &79.8 \\
\scratch MinkUNet~\cite{choy20194d} &2019 &65.4 &65.4 \\
\pretrain\ + PC~\cite{xie2020pointcontrast} &2020 &70.3 &- \\
\pretrain\ + CSC~\cite{hou2021exploring} &2021 &72.2 &- \\
\pretrain\ + MSC~\cite{wu2023masked} &2023 &70.1 &- \\
\pretrain\ + GC~\cite{wang2024gc} &2024 &72.0 &- \\
\cellcolor[HTML]{efefef}\pretrain\ + PPT~(Ours) &\cellcolor[HTML]{efefef}2024 &\cellcolor[HTML]{efefef}72.7 &\cellcolor[HTML]{efefef}78.1 \\
\scratch PTv3~\cite{pointcept2023} &2024 &73.4 &77.7 \\
\cellcolor[HTML]{efefef}\pretrain\ + PPT~(Ours) &\cellcolor[HTML]{efefef}2024 &\cellcolor[HTML]{efefef}\textbf{74.7} &\cellcolor[HTML]{efefef}\textbf{80.8} \\
\bottomrule
\end{tabular}

        \vspace{-2mm}
        \caption{\textbf{S3DIS semantic segmentation.} }\label{tab:appendix_s3dis}
        \vspace{-6mm}
    \end{minipage}
\end{table}

\addcontentsline{toc}{subsubsection}{(3)\ \ \ \ LCA as Prediction Head}
\mypara{LCA as prediction head.}
We introduce Language-guided Categorical Alignment (LCA) as a method to align the category spaces across multiple datasets with a unified category-text embedding. This alignment strategy can also be employed as a segmentation prediction head within a standard single dataset training process. By considering the scaled similarity between point embedding and category-text embedding as the predicted logit value, LCA serves as an effective prediction head. In \tabref{subtab:lca_head}, we compare the performance of the standard linear prediction head with LCA as the prediction head. The experimental results demonstrate that LCA can also enhance model performance in the context of standard single dataset segmentation tasks.

\section{Additional Comparision}
In this section, we expand upon the combined results table for semantic segmentation (\tabref{tab:sem_seg} and \tabref{tab:sem_seg_outdoor}) from our main paper, offering a more detailed breakdown of results alongside the respective publication years of previous works. This comprehensive result table is designed to assist readers in tracking the progression of research efforts in 3D representation learning. Marker \scratch refers to the result from a model trained from scratch, and \pretrain refers to the result from a pre-trained model.

\begin{table}[!t]
    \begin{minipage}{0.48\textwidth}
    \centering
        \tablestyle{15pt}{1.08}
        \begin{tabular}{lccc}\toprule
Methods &Year &Val &Test \\\midrule
\scratch SPVNAS~\cite{tang2020spvnas} &2020 &64.7 &66.4 \\
\scratch Cylinder3D~\cite{zhu2021cylindrical} &2021 &64.3 &67.8 \\
\scratch PVKD~\cite{hou2022pvkd} &2022 &- &71.2 \\
\scratch 2DPASS~\cite{yan20222dpass} &2022 &69.3 &72.9 \\
\scratch PTv2~\cite{wu2022point} &2022 &70.3 &72.6 \\
\scratch WaffleIron~\cite{puy23waffleiron} &2023 &68.0 &70.8 \\
\scratch SphereFormer~\cite{lai2023spherical} &2023 &67.8 &74.8 \\
\scratch RangeFormer~\cite{kong2023rangeformer} &2023 &67.6 &73.3 \\
\scratch MinkUNet~\cite{choy20194d} &2019 &63.8 &- \\
\cellcolor[HTML]{efefef}\pretrain\ + PPT~(Ours) &\cellcolor[HTML]{efefef}2024 &\cellcolor[HTML]{efefef}71.4 &\cellcolor[HTML]{efefef}- \\
\scratch OA-CNNs~\cite{peng2024oacnns} &2024 &70.6 &- \\
\scratch PTv3~\cite{pointcept2023} &2024 &70.8 &74.2 \\
\pretrain\ + M3Net~\cite{liu2024m3net} &2024 &72.0 &75.1 \\
\cellcolor[HTML]{efefef}\pretrain\ + PPT~(Ours) &\cellcolor[HTML]{efefef}2024 &\cellcolor[HTML]{efefef}\textbf{72.3} &\cellcolor[HTML]{efefef}\textbf{75.5} \\
\bottomrule
\end{tabular}
        \vspace{-2mm}
        \caption{\textbf{SemanticKITTI semantic segmentation.} }\label{tab:appendix_semantic_kitti}
        \vspace{2mm}
    \end{minipage}
    \begin{minipage}{0.48\textwidth}
    \centering
        \tablestyle{15pt}{1.08}
        \begin{tabular}{lrrrr}\toprule
Methods &Year &Val &Test \\\midrule
\scratch SPVNAS~\cite{tang2020spvnas} &2020 &77.4 &- \\
\scratch Cylinder3D~\cite{zhu2021cylindrical} &2021 &76.1 &77.2 \\
\scratch PVKD~\cite{hou2022pvkd} &2022 &- &76.0 \\
\scratch 2DPASS~\cite{yan20222dpass} &2022 &- &80.8 \\
\scratch PTv2~\cite{wu2022point} &2022 &80.2 &82.6 \\
\scratch SphereFormer~\cite{lai2023spherical} &2023 &78.4 &81.9 \\
\scratch RangeFormer~\cite{kong2023rangeformer} &2023 &78.1 &80.1 \\
\scratch MinkUNet~\cite{choy20194d} &2019 &73.3 &- \\
\cellcolor[HTML]{efefef}\pretrain\ + PPT~(Ours) &\cellcolor[HTML]{efefef}2024 &\cellcolor[HTML]{efefef}78.6 \cellcolor[HTML]{efefef}&\cellcolor[HTML]{efefef}- \\
\scratch OA-CNNs~\cite{peng2024oacnns} &2024 &78.9 &- \\
\scratch PTv3~\cite{pointcept2023} &2024 &80.4 &82.7 \\
\cellcolor[HTML]{efefef}\pretrain\ + PPT~(Ours) &\cellcolor[HTML]{efefef}2024 &\cellcolor[HTML]{efefef}\textbf{81.2} &\cellcolor[HTML]{efefef}\textbf{83.0} \\
\bottomrule
\end{tabular}
        \vspace{-2mm}
        \caption{\textbf{NuScenes semantic segmentation.} }\label{tab:appendix_nuscenes}
        \vspace{-6mm}
    \end{minipage}
\end{table}

\subsection{Indoor Semantic Segmentation}

We conduct a detailed comparison of pre-training technologies and backbones on the ScanNet v2~\cite{dai2017scannet} (see \tabref{tab:appendix_scannet}) and S3DIS~\cite{armeni2016s3dis} (see \tabref{tab:appendix_s3dis}) datasets. ScanNet v2 comprises 1,513 room scans reconstructed from RGB-D frames, divided into 1,201 training scenes and 312 for validation. In this dataset, model input point clouds are sampled from the vertices of reconstructed meshes, with each point assigned a semantic label from 20 categories (e.g., wall, floor, table). The S3DIS dataset for semantic scene parsing includes 271 rooms across six areas from three buildings. Following a common practice~\cite{tchapmi2017segcloud,qi2017pointnet++,zhao2021point}, we withhold area 5 for testing and perform a 6-fold cross-validation. Different from ScanNet v2, S3DIS densely sampled points on mesh surfaces, annotated into 13 categories. Consistent with standard practice~\citep{qi2017pointnet++}. We employ the mean class-wise intersection over union (mIoU) as the primary evaluation metric for indoor semantic segmentation.

\subsection{Outdoor Semantic Segmentation}
We extend our comprehensive evaluation of pre-training technologies and backbones to outdoor semantic segmentation tasks, focusing on the SemanticKITTI~\cite{behley2019semantickitti}(see \tabref{tab:appendix_semantic_kitti}) and NuScenes~\cite{caesar2020nuscenes} (see \tabref{tab:appendix_nuscenes}) datasets. SemanticKITTI is derived from the KITTI Vision Benchmark Suite and consists of 22 sequences, with 19 for training and the remaining 3 for testing. It features richly annotated LiDAR scans, offering a diverse array of driving scenarios. Each point in this dataset is labeled with one of 28 semantic classes, encompassing various elements of urban driving environments. NuScenes, on the other hand, provides a large-scale dataset for autonomous driving, comprising 1,000 diverse urban driving scenes from Boston and Singapore. For outdoor semantic segmentation, we also employ the mean class-wise intersection over union (mIoU) as the primary evaluation metric for outdoor semantic segmentation.

{
\small
\bibliographystyle{ieeenat_fullname}
\bibliography{main}
}

\end{document}